\documentclass{article}
\usepackage{iclr2023_conference,times}

\usepackage{amsmath,amsfonts,bm}

\def\eqref#1{equation~\ref{#1}}

\def\1{\bm{1}}

\DeclareMathAlphabet{\mathsfit}{\encodingdefault}{\sfdefault}{m}{sl}
\SetMathAlphabet{\mathsfit}{bold}{\encodingdefault}{\sfdefault}{bx}{n}

\newcommand{\E}{\mathbb{E}}

\newcommand{\KL}{D_{\mathrm{KL}}}

\usepackage[pagebackref,breaklinks,colorlinks]{hyperref}
\usepackage{url}

\usepackage{algorithm}
\usepackage{adjustbox}
\usepackage{url}
\usepackage{epigraph}
\usepackage{booktabs}
\usepackage{bbding}
\usepackage{listings}
\usepackage{tablefootnote}
\usepackage{wrapfig}
\usepackage{arydshln}
\usepackage{subcaption}
\usepackage{multicol}

\usepackage{xcolor}
\definecolor{green}{rgb}{0.0, 0.42, 0.24} 
\definecolor{orange}{rgb}{0.8, 0.33, 0.} 
\definecolor{blue}{rgb}{0.16, 0.32, 0.75} 
\definecolor{cobalt}{rgb}{0.0, 0.28, 0.67} 
\definecolor{egred}{rgb}{1.0, 0.25, 0.25}

\definecolor{codegreen}{rgb}{0,0.6,0}
\definecolor{codegray}{rgb}{0.5,0.5,0.5}
\definecolor{codepurple}{rgb}{0.58,0,0.82}
\definecolor{backcolour}{rgb}{0.95,0.95,0.92}

\definecolor{relic_color}{HTML}{2C7BB6}

\lstdefinestyle{mystyle}{
    backgroundcolor=\color{backcolour},   
    commentstyle=\color{codegreen},
    keywordstyle=\color{magenta},
    numberstyle=\tiny\color{codegray},
    stringstyle=\color{codepurple},
    basicstyle=\ttfamily\scriptsize,
    breakatwhitespace=false,         
    breaklines=true,                 
    captionpos=b,
    escapeinside={\%*}{*)},
    keepspaces=true,                 
    numbers=left,                    
    numbersep=5pt,                  
    showspaces=false,                
    showstringspaces=false,
    showtabs=false,                  
    tabsize=2
}

\setcitestyle{square}

\newcommand{\relic}{\textsc{ReLIC}}
\newcommand{\alpharelic}{\textsc{ReLIC}v2}

\usepackage[capitalize]{cleveref}
\crefname{section}{Sec.}{Secs.}
\Crefname{section}{Section}{Sections}
\Crefname{table}{Table}{Tables}
\crefname{table}{Tab.}{Tabs.}

\iclrfinalcopy 

\begin{document}

\begingroup\centering

\title{Pushing the limits of self-supervised ResNets:\\Can we outperform supervised learning without labels on ImageNet?}

\author{Nenad Tomasev\textsuperscript{\textsection} \\
DeepMind \\ London, UK 
\and
{\bf Ioana Bica\textsuperscript{\textsection}} \\ 
DeepMind \\ London, UK 
\and 
{\bf Brian McWilliams\textsuperscript{\textsection, *}} \\
Twitter Cortex \\
\and 
{\bf Lars Buesing} \\
DeepMind \\ London, UK \\
\and 
{\bf Razvan Pascanu} \\
DeepMind \\ London, UK \\
\and 
{\bf Charles Blundell} \\
DeepMind \\ London, UK \\
\and 
{\bf Jovana Mitrovic\textsuperscript{\textsection,$\dagger$}} \\
DeepMind \\ London, UK \\
}

\begingroup\renewcommand\thefootnote{\textsection}
\footnotetext{Equal contribution}
\endgroup

\begingroup\renewcommand\thefootnote{$\dagger$}
\footnotetext{Primary contact: mitrovic@google.com}
\endgroup

\begingroup\renewcommand\thefootnote{*}
\footnotetext{Work done while at DeepMind.}
\endgroup

\maketitle

\endgroup

\begin{abstract}
Despite recent progress made by self-supervised methods in representation learning with residual networks, they still underperform supervised learning on the ImageNet classification benchmark, limiting their applicability in performance-critical settings.
Building on prior theoretical insights from \relic{} \citep{mitrovic2020representation}, we include additional inductive biases into self-supervised learning.
We propose a new self-supervised representation learning method, \alpharelic{}, which combines an explicit invariance loss with a contrastive objective over a varied set of appropriately constructed data views to avoid learning spurious correlations and obtain more informative representations.
\alpharelic{} achieves $77.1\%$ top-$1$ accuracy on ImageNet under linear evaluation on a ResNet50, thus improving the previous state-of-the-art by absolute $+1.5\%$; on larger ResNet models, \alpharelic{} achieves up to $80.6\%$ outperforming previous self-supervised approaches with margins up to $+2.3\%$.
Most notably, \alpharelic{} is the first unsupervised representation learning method to consistently outperform the supervised baseline in a like-for-like comparison over a range of ResNet architectures.
Using \alpharelic{}, we also learn more robust and transferable representations that generalize better out-of-distribution than previous work, both on image classification and semantic segmentation.
Finally, we show that despite using ResNet encoders, \alpharelic{} is comparable to state-of-the-art self-supervised vision transformers.
\end{abstract}

\section{Introduction}
\label{sec:intro}

Large-scale \emph{foundation models} \citep{bommasani2021opportunities}---in particular for language \citep{devlin2018bert, brown2020language} and multimodal domains \citep{radford2021learning}---are an important recent development in representation learning. The idea that massive models can be trained without labels in an unsupervised (or self-supervised) manner and be readily adapted, in a few- or zero-shot setting, to perform well on a variety of downstream tasks is important for many problem areas for which labeled data is expensive or impractical to obtain. 
\emph{Contrastive} objectives have emerged as a successful strategy for representation learning \citep{chen2020simple, he2019momentum, mitrovic2020representation}. However, downstream utility\footnote{Downstream utility is commonly measured by how well a method performs under the standard linear evaluation protocol on ImageNet; see section~\ref{sec:results:linear}.} of these representations has until now never exceeded the performance of supervised training of the same architecture, thus limiting their wide adoption.

\begin{wrapfigure}{R}{0.5\textwidth}
    \centering
    \includegraphics[width=0.5\textwidth]{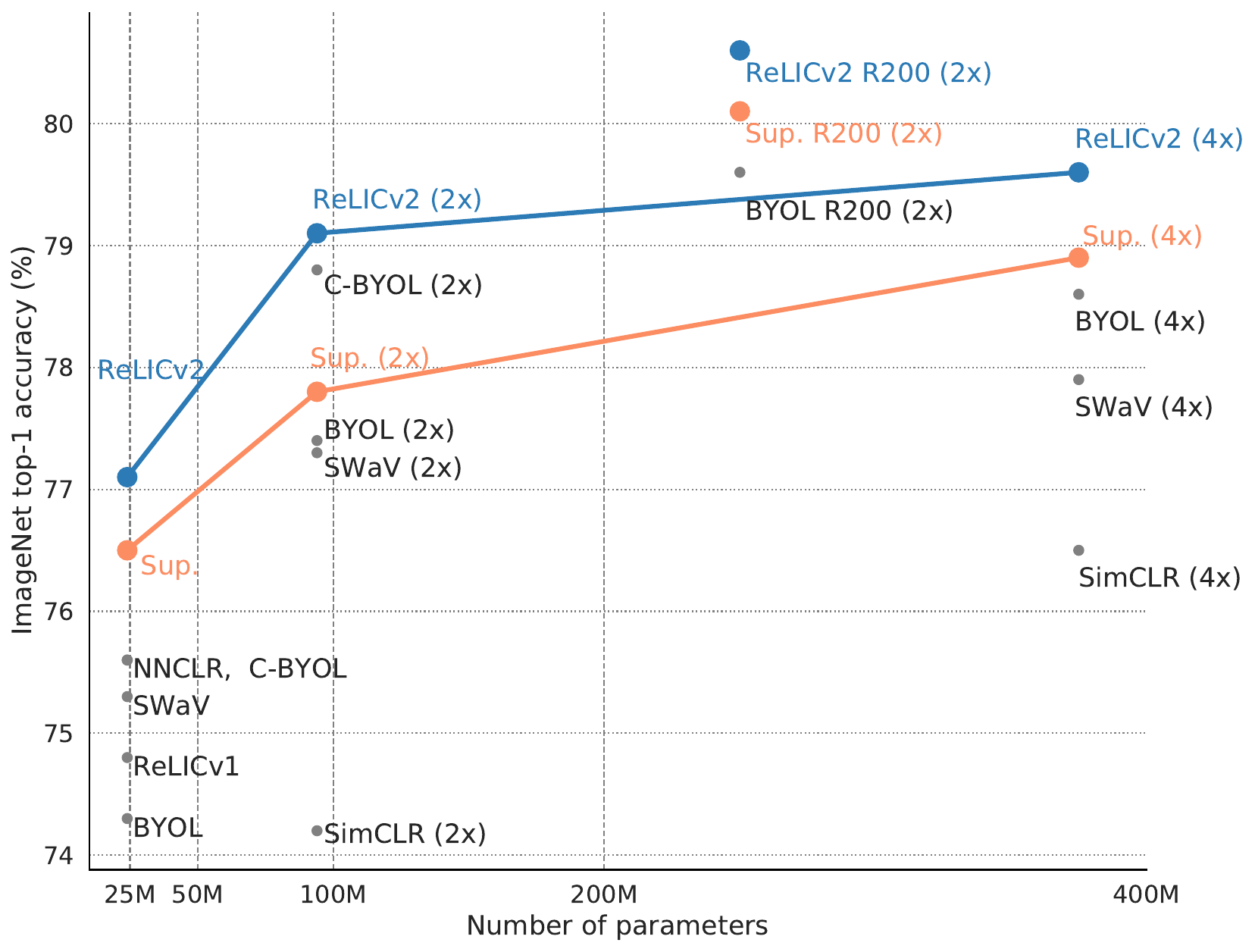}
  \caption{Top-1 linear evaluation accuracy on ImageNet using ResNet50 encoders with $1\times$, $2\times$ and $4\times$ width multipliers and a ResNet200 encoder with a $2\times$ width multiplier.
  \label{fig:top1}}
  \vspace{-0.5cm}
\end{wrapfigure}

In this work, we tackle the question ``Can we outperform supervised learning without labels on ImageNet?". Thus, our focus is on learning good representations for high-level vision tasks such as image classification.
In supervised learning we have access to label information which provides a signal for what features are relevant for classification.
In unsupervised learning there is no such signal and we have strictly less information available from which to learn compared to the supervised setting.
Thus, the challenge of outperforming supervised learning without labels might seem impossible.
Comparing supervised and contrastive objectives we see that they are two very different approaches to learning representations that yield significantly different representations.
While supervised approaches use labels as targets within a cross-entropy objective, contrastive methods rely on comparing datapoints against other similar and dissimilar datapoints.
Thus, supervised representations end up encoding a small set of highly informative features for downstream performance, while contrastive representations encode many more features with some of these features not related to downstream performance. 
This intuition is also supported by the observation that when visualizing the encoded information of contrastive representations through reconstruction, they are found to retain more detailed information of the original image, such as background and style, than supervised representations \citep{pbordes2021high}. 

Based on this, we hypothesize that one of the key reasons for the current subpar performance of contrastive (and thus self-supervised) representations, is the presence of features which are not directly related to downstream tasks, i.e. so-called \emph{spurious features}.
In general, basing representations on spurious features can have negative consequences for the model's generalization performance and thus avoiding to encode these features is paramount for learning informative representations.

In this paper, we propose to equip self-supervised methods with additional inductive biases to obtain more informative representations and overcome the lack of additional information that supervised methods have access to.
We use as our base self-supervised approach the performant \relic{} method \citep{mitrovic2020representation} which combines a contrastive loss with an invariance loss.
We propose to extend \relic{} by adding inductive biases that penalize the learning of spurious features such as background and style; we denote this method as \alpharelic{}.
First, we propose a new fully unsupervised saliency masking pipeline which enables us to separate the image foreground from its background. We include this novel saliency masking approach as part of the data augmentation pipeline.
Leveraging the invariance loss, this enables us to learn representations that do not rely on the spurious background features.
Second, while \relic{} operates on just two data views of the same size, we extend \alpharelic{} to multiple data views of varying sizes.
Specifically, we argue for learning from a large number of data views of the whole image as well as including a small number of data views that encode only a small part of the image.
The intuition for using smaller crops is that this enables learning of more robust representations as not all features of the foreground might be contained in the small crop. 
Thus, the learned representation is robust against individual object features being absent as it is the case in many real-world settings, e.g. only parts of an object are visible because of occlusion.
On the other hand, using multiple large crops enables us to learn representations that are invariant to object style as the different views correspond to different augmentations.
We extend the contrastive and invariance losses of \relic{} to operate over multiple views of varying sizes.

To showcase the generality and wide applicability of \alpharelic{}, we test it on a wide range of downstream tasks and datasets including image classification, semi-supervised and transfer learning, robustness and out-of-distribution generalisation.
First, we show that \alpharelic{} outperforms by a significant margin previous state-of-the-art self-supervised methods on a wide range of ResNet architectures (ResNet50, ResNet 50 2x, ResNet 50 4x, ResNet101, ResNet152, ResNet200, ResNet 200 2x).  
In terms of top-$1$ classification accuracy on ImageNet, \alpharelic{} achieves $77.1\%$ with ResNet50 which is a $+1.5\%$ improvement over previous work; on larger ResNets, specifically with ResNet200 $2\times$, \alpharelic{} achieves $80.6\%$ top-$1$ accuracy.
Second, we show that \alpharelic{} learns more transferable and robust representations that generalize better out of distribution than previous methods both on image classification as well as semantic segmentation tasks.
Finally, as shown in Figure~\ref{fig:top1}, \alpharelic{} is the first self-supervised representation learning method that outperforms a standard supervised baseline on linear ImageNet evaluation across a wide range of ResNet architectures in a like-for-like comparison.\footnote{Concurrent work in \citep{lee2021compressive} outperforms the same standard supervised baseline only on a ResNet50 $2\times$ encoder.}
Although using ResNets, \alpharelic{} demonstrates comparable performance to recent Vision Transformers \citep{dosovitskiy2020image} (Figure~\ref{fig:top1_vit}). 

These strong experimental results across different vision tasks, learning settings and datasets showcase the generality of our proposed representation learning method. Moreover, we believe that the concepts and results developed in this work could have important implications for wider adoption of self-supervised pre-training in a variety of domains as well as the design of objectives for foundational machine learning systems.

\noindent\textbf{Summary of contributions.}
We tackle the question “Can we outperform supervised learning without labels on ImageNet?”.
We propose to extend the self-supervised method \relic{} \citep{mitrovic2020representation} with additional inductive biases to learn more informative representations.
We develop a fully unsupervised saliency masking pipeline and use multiple data views of varying sizes to encode these inductive biases. 
The resulting method \alpharelic{} significantly outperforms previous state-of-the-art self-supervised methods across a wide range of ResNet architectures of different depths and width. 
We also highlight the generality of \alpharelic{} representations through state-of-the-art performance on transfer learning, semi-supervised learning, and robustness and out-of-distribution generalization, thus outperforming competing self-supervised methods by a wide margin.
Furthermore, \alpharelic{} is the first self-supervised representation learning method that outperforms a standard supervised model across a wide range of ResNet architecture of different depths and widths.

\section{Method} \label{sec:method}

\subsection{Background}

Representation Learning via Invariant Causal Mechanisms (\relic{}) \citep{mitrovic2020representation} learns representations by comparing two differently augmented data views through two mechanisms -- \textit{instance classification}
and \textit{invariant prediction}.
As is common in prior work, \relic{} tackles the instance classification problem with a contrastive objective, i.e. it learns representations by maximizing the similarity between two differently augmented views of the same datapoint (positives), while minimizing the similarity of that datapoint with other datapoints (negatives).
In addition to that, \relic{} introduces an invariance objective which ensures that the distribution of similarities of a datapoint with its positives and negatives is invariant across augmentations of that datapoint.

Given a randomly sampled batch of datapoints $\{x_{i}\}_{i=1}^{N}$ with $N$ the batch size, \relic{} learns an encoder $f$ that outputs the representation $z$, i.e. $z_{i} = f(x_{i})$. Following \citep{chen2020simple, grill2020bootstrap}, \relic{} creates two views of the data by applying two distinct augmentations randomly sampled from the data augmentation pipeline proposed in \citep{chen2020simple}, i.e. $t, t^{\prime} \sim \mathcal{T}$; this yields two augmented batches $\{x_{i}^{t}\}_{i=1}^{N}$ and $\{x_{i}^{t^{\prime}}\}_{i=1}^{N}$.  
To solve the instance classification problem, \relic{} maximizes the following probability
\begin{equation}
    p(x_i^{t}; x_i^{t^{\prime}})
    = \frac{
     e^{\phi_\tau\left(x_i^{t}, x_i^{t^{\prime}}\right)}
    }{
    e^{\phi_\tau\left( x_i^t, x_i^{t^{\prime}}\right)} + \sum_{x_{j}^{t^\prime} \in \mathcal{N}(x_{i})}  e^{\phi_\tau\left(x_i^{t}, x_j^{t^{\prime}}\right)}
    }
\label{eq:likelihood}
\end{equation}
where $\phi_\tau(x_i, x_j)= \langle h(f(x_i)), q(g(x_j)) \rangle/\tau$ measures the similarity between representations with $\tau$ the temperature parameter.
\relic{} adopts the \emph{target} network setting of \citep{grill2020bootstrap} such that $f$ and $g$ have the same architecture, but the weights of $g$ are an exponential moving average of the weights of $f$; also, $h$ and $q$ are multi-layer perceptrons with $h$ playing the role of the composition of the projector and predictor from \citep{grill2020bootstrap} and $q$ being the exponential moving average of the projector network.; we refer to $f$ and $h$ taken together as the online network, while $g$ and $q$ comprise the target network. 
$\mathcal{N}(x_i)$ represents the set of \emph{negatives}; we uniformly randomly sample a small number of points from the batch to serve as negatives following \citep{mitrovic2020less}.

In order to optimize for invariant prediction, \relic{} introduces an \emph{invariance loss} defined as the Kullback-Leibler divergence between the likelihood of the two augmented data views as 
\begin{equation}
 \KL (p(x_i^{t}) \parallel p(x_i^{t^{\prime}})) = \text{sg}\left[\E_{p(x_i^{t}; x_i^{t^{\prime}})} \log p(x_i^{t}; x_i^{t^{\prime}})\right] 
 -  \E_{p(x_i^{t}; x_i^{t^{\prime}})}\log p(x_i^{t^{\prime}}; x_i^{t})
  \label{eq:invariance}
\end{equation}
and we use the shorthands $p(x_i^{t}) = p(x_i^{t}; x_i^{t^{\prime}})$ and $p(x_i^{t^\prime}) = p(x_i^{t^\prime}; x_i^{t})$.
The invariance loss enforces that the similarity of the representations $ f(x_i^{t})$ and $ f(x_i^{t^\prime})$ \emph{relative} to the positives and negatives is the same; the stop gradient operator, $\text{sg}[\cdot]$, does not affect the computation of the KL-divergence but avoids degenerate solutions during optimization \citep{xie2019unsupervised}.
Taken together, \relic{} learns representations by optimizing the following loss
\begin{equation}
    \mathcal{L} = \sum_{t, t^{\prime}\sim\mathcal{T}} \sum_{i=1}^{N} -\log p(x_i^{t}; x_i^{t^{\prime}}) + \beta\KL (p(x_i^{t}) \parallel p(x_i^{t^{\prime}}))
    \label{eq:loss_relic}
\end{equation}
with $\beta$ a scalar weighting the relative importance of the contrastive and invariance losses.

\subsection{\alpharelic{}} 
\begin{figure}[t]
    \centering
    \includegraphics[width=1.0\textwidth]{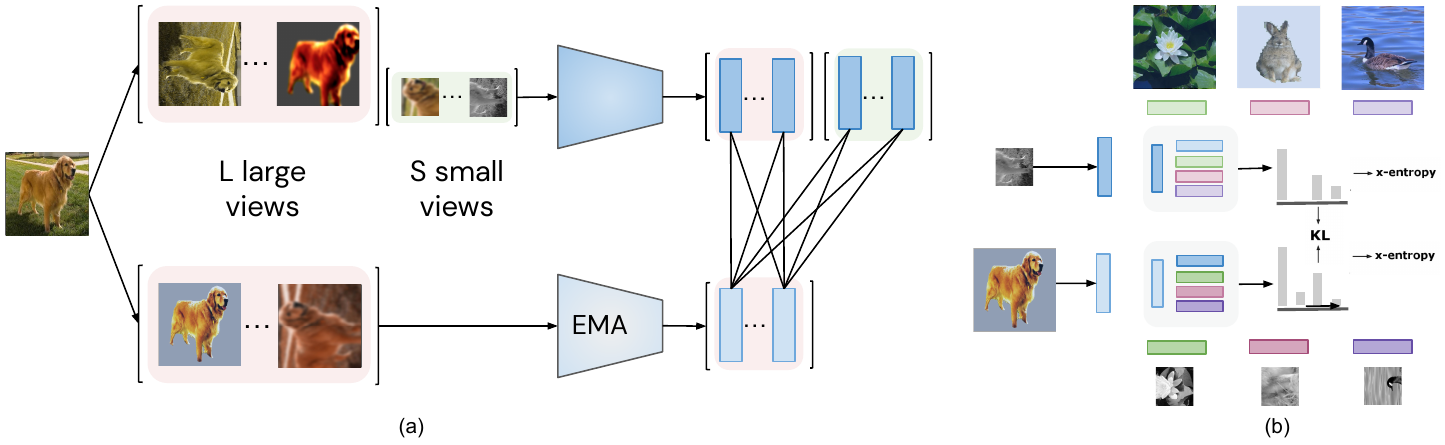}
  \caption{(a) \alpharelic{} uses saliency masking as part of the data augmentations pipeline and views of various sizes to learn representations that are invariant to spurious correlations. Note that the $L$ (differently augmented) large views are passed through both the online and target networks, while the small $S$ views are only passed through the online network. The learning objective is computed by comparing each of the large and small views passed through the online network with each large view passed through the target network. (b)  The objective used for each comparison combines the contrastive (instance discrimination) loss, i.e the cross-entropy (x-entropy) loss based on the similarity scores, and the invariance loss, i.e. the Kullback-Leibler (KL) divergence between the similarity scores across augmentations. 
\label{fig:relicv2}}
\vspace{-0.5cm}
\end{figure}

We extend Representation Learning via Invariant Causal Mechanisms (\relic{}) \citep{mitrovic2020representation} to include additional inductive biases that prevent the learning of spurious features which are hurting the learning of informative representations.
We encode these inductive biases through a novel fully unsupervised saliency masking pipeline as well as scaling up \relic{} to accommodate multiple data views of varying sizes; we call this method \alpharelic{} and visualize it in Figure \ref{fig:relicv2}.

\noindent\textbf{Saliency masking.} To localize the semantically relevant parts of the image, we propose to use saliency masking. 
In order to separate foreground from background, we develop a fully unsupervised saliency masking pipeline that does not use any additional data.
We learn a saliency masking network without access to any labels or any additional data outside of the ImageNet training set which we also use for pre-training the representation.
We start by using a small subset of ImageNet ($2500$ randomly selected images) and compute initial estimates of saliency masks using a number of handcrafted methods.
Specifically, we use Robust Background Detection  \citep{zhu2014saliency}, Manifold Ranking  \citep{yang2013saliency}, Dense and Sparse Reconstruction  \citep{li2013saliency} and Markov Chain  \citep{jiang2013saliency}. 
Next, we use a ResNet50 2x network pretrained on ImageNet in a self-supervised way as the initial saliency detection network.
We adopt the self-supervised refinement mechanism from DeepUSPS\footnote{Note that DeepUSPS cannot be directly applied in our setting as it relies on labelled information and additional datasets (CityScapes) \citep{Cordts2016Cityscapes}.} \citep{deepusps} and incrementally refine the handcrafted saliency masks using a self-supervised objective. Subsequently, the saliency detection network is trained by fusing the refined handcrafted saliency masks.
The trained saliency detection network is used for computing the saliency masks in our pipeline.
During \alpharelic{} pre-training, we randomly apply the saliency masking to input images with a small probability $p_m$. By leveraging images where the background has been removed, \alpharelic{} learns representations that focus on foreground features and are robust to background changes as well as any spurious features that might be found in the background. For more details see  appendix.

\noindent\textbf{Views of varying sizes.}
We propose to enforce invariance over multiple randomly augmented data views of varying sizes to learn more informative representations.
To achieve this, we use a large number of views encoding the whole image as well as a small number of smaller views which contain only a portion of the image.\footnote{Most other methods use only $2$ data views of the whole image.}
As the number of views of the whole image increases so does the variation of object styles seen during training as each view represents a random augmentation manipulating the style of the image. 
Thus, explicitly enforcing invariance over a diverse set of object styles enables us to learn representations which are increasingly invariant to spurious changes in object style.  
Incorporating small views which are random crops of the original image into the learning serves two purposes. 
First, as these views represent a small part of the original image, it is likely that some parts of the objects of interest might be occluded.
Thus, as we learn representations through invariance, using these small views enables us to learn representations which are more robust to object occlusions, a common issue in real-world data.
Note that it is important to use only a small number of small views (in comparison to the number of large views) as using too many small views can result in representations that fail to encode important features as different features might be occluded in different small views and we are learning representations by enforcing invariance.\footnote{For this purpose, we use only $2$ small views in our experiments and provide an ablation analysis on the different number of large and small views.}
Second, we hypothesize that small crops play a synergistic role to saliency masking as taking a small crop of the image is likely to remove potentially large parts of the background; see analysis section for some experimental validation.\footnote{Note that saliency masking is not perfect at removing the background and struggles especially in settings where there is little color contrast between the background and foreground.}

\noindent\textbf{Objective.}
\alpharelic{} optimizes a combination of the contrastive and invariance losses across  differently augmented large and small views.
In particular, \alpharelic{} optimizes the following objective:
\begin{align}
    \mathcal{L} = \sum_{i=1}^{N} \sum_{1 \leq l_{1}\leq L}  &
     \biggl( \sum_{1 \leq l_{2} \leq L} \left( 
        -\log p(x_i^{l, t_{l_2}}; x_i^{l, t_{l_{1}}}) +
         \beta\KL (p(x_i^{l, t_{l_{2}}})  \parallel p(x_i^{l, t_{l_{1}}})) \right)+ \\ \nonumber
 & \sum_{1\leq s_{1} \leq S} \left(-\log p(x_i^{s, t_{s_{1}}}; x_i^{l, t_{l_{1}}})  + \beta\KL (p(x_i^{s, t_{s_{1}}}) \parallel p(x_i^{l, t_{l_{1}}}))\right)\biggr)
\end{align}
with $x_{i}^{l, t_{l}}$ a large size view under augmentation $t_{l} \sim \mathcal{T}_{\text{sal}}$, $x_{i}^{s, t_s}$ a small size view under augmentation $t_{s} \sim \mathcal{T}$ and $L$ and $S$ the number of large and small crops, respectively; where $\mathcal{T}_{\text{sal}}$ adds saliency masking with probability $p_m$ on top of $\mathcal{T}$. For more details refer to the appendix. Note that we leverage the differently sized data views in different ways for learning representations. 
We use the large views both for updating the online network as well as for computing learning targets through the target network, i.e. $x_{i}^{l, t_{l_\cdot}}$ appears on both sides of $p$. 
On the other hand, we only use the small views for updating the online network and not as learning targets, i.e. $x_{i}^{s, t_{s_1}}$ appears only on the left hand side of $p$, c.f. Equation \ref{eq:likelihood}. 
For the precise architectural and implementation details, as well as a pseudo-code for \alpharelic{} see the appendix.

\section{Evaluating pretrained representations on ImageNet}
\label{sec:results:linear}

We use \alpharelic{} to pretrain representations on the training set of the ImageNet ILSVRC-2012 dataset \citep{russakovsky2015imagenet} without using labels. We evaluate \alpharelic{} representations by training a linear classifier on top of the frozen representation according to the procedure described in \citep{chen2020simple,grill2020bootstrap,dwibedi2021little} and the appendix. From Table~\ref{table.imagenet_linear}, for the ResNet50 encoder, we see that \alpharelic{}  outperforms all previous self-supervised approaches by a significant margin of $+1.5\%$ in terms of top-1 accuracy on the ImageNet test set. \begin{wraptable}{r}{0.4\textwidth}
\vspace{-0.2cm}
\begin{center}
\caption{Top-1 accuracy (in \%) under linear evaluation on the ImageNet test set for a ResNet50 encoder set for different representation learning methods.
}
\label{table.imagenet_linear}
\begin{adjustbox}{max width=0.4\textwidth}
\begin{tabular}{lcc}
\toprule
Method & Top-1 \\
\toprule
\; Supervised \tiny\citep{chen2020simple}   & 76.5 \\ 
\midrule
\; SimCLR {\tiny\citep{chen2020simple}}  & 69.3 \\
\; MoCo v2 \tiny\citep{Chen2020ImprovedBW} & 71.1 \\
\; InfoMin Aug. \tiny\citep{Tian2020WhatMF} & 73.0  \\
\; BYOL \tiny\citep{grill2020bootstrap} & 74.3  \\
\; \relic{} \tiny\citep{mitrovic2020representation} & 74.8 \\ 
\; SwAV \tiny\citep{caron2020unsupervised} & 75.3  \\
\; NNCLR  \tiny\citep{dwibedi2021little}  & 75.6  \\
\; C-BYOL \tiny\citep{lee2021compressive} & 75.6  \\
\; \alpharelic{} (ours) & \textbf{77.1} \\
\hline
\end{tabular}
\end{adjustbox}
\end{center}
\vspace{-0.3cm}
\end{wraptable}
 Moreover, in Table \ref{tab:linear_other_resnets}, note that \alpharelic{} achieves state-of-the-art performance, with margins up to $+2.3\%$ in absolute terms, across a varied set of ResNet encoders of different sizes, spanning ResNet50, ResNet101, ResNet152 and ResNet200 and layer widths of $1\times$, $2\times$ and $4\times$.
ResNet50 with $2\times$ and $4\times$ wider layers has $94$ and $375$ million parameters, respectively.
ResNet101, ResNet152, ResNet200 and ResNet200 $2\times$ have $43$, $58$, $63$ and $250$ million parameters, respectively. Please refer to the appendix for additional results and more details.

Remarkably, from Tables \ref{table.imagenet_linear} and \ref{tab:linear_other_resnets}, we also notice that \alpharelic{} outperforms the standard supervised baselines, across a wide range of ResNet architectures, despite using no label information to pretrain the representation. 
For the ResNet50 encoder, we compare against the supervised baseline model used through the representation learning literature (c.f.~\citep{chen2020simple,grill2020bootstrap,caron2020unsupervised,dwibedi2021little}). 
This model is trained with a cross-entropy loss, a cosine learning rate schedule and full access to labels using the same set of data augmentations as proposed by \citep{chen2020simple}. 
While the recent work of \citep{wightman2021resnet} proposes a series of elaborate optimization and data augmentation tricks to improve the performance of the supervised ResNet50 on ImageNet, we leave the application of these heuristics for future work.
Instead, in this work, we focus on a fair like-for-like comparison between self-supervised methods and supervised learning 
 \footnote{Note that using saliency masking to create background augmentations for training the supervised baseline does not improve performance as highlighted in \cite{ryali2021leveraging}}.
Refer to the appendix for details about the supervised baselines we used for the other ResNet encoders.

\begin{table}[h]
\begin{minipage}{.5\linewidth}
    \centering
\begin{tabular}{lc}
\toprule
Method & Top-1  \\
\midrule
\; Supervised \tiny\citep{chen2020simple} &  77.8  \\ 
\; MoCo \tiny\citep{he2019momentum} & 65.4  \\
\; SimCLR \tiny\citep{chen2020simple} & 74.2  \\
\; BYOL \tiny\citep{grill2020bootstrap} &  77.4  \\
\; SwAV \tiny\citep{caron2020unsupervised} &  77.3  \\
\; C-BYOL \tiny\citep{lee2021compressive}  & 78.8  \\
\; \alpharelic{} (ours) &   {\bf 79.0}  \\
\bottomrule
\end{tabular}
\subcaption{ResNet50 $2\times$ encoder.}
\end{minipage}%
    \begin{minipage}{.5\linewidth}
\centering
\begin{tabular}{lc}
\toprule
Method & Top-1  \\
\midrule
\; Supervised \tiny\citep{chen2020simple} &  78.9  \\ 
\; MoCo \tiny\citep{he2019momentum} & 68.6  \\
\; SimCLR \tiny\citep{chen2020simple}  & 76.5 \\
\; SwAV \tiny\citep{caron2020unsupervised} &  77.9  \\
\; BYOL \tiny\citep{grill2020bootstrap} &  78.6  \\
\; \alpharelic{} (ours) &  {\bf 79.4} \\  
\bottomrule
\end{tabular}
\subcaption{ResNet50 $4\times$ encoder.}
\end{minipage}
\begin{minipage}{0.5\linewidth}
\centering
\begin{tabular}{lc}
\toprule
Method & Top-1  \\
\midrule
\; Supervised \tiny\citep{grill2020bootstrap} & 78.0  \\ 
\; BYOL \tiny\citep{grill2020bootstrap} & 76.4  \\
\; \alpharelic{} (ours) & {\bf 78.7}   \\ 
\bottomrule
\end{tabular}
\subcaption{ResNet101 encoder.}
\end{minipage}%
\begin{minipage}{0.5\linewidth}
\centering
\begin{tabular}{lc}
\toprule
Method & Top-1  \\
\midrule
\; Supervised \tiny\citep{grill2020bootstrap} & 79.1  \\ 
\; BYOL \tiny\citep{grill2020bootstrap} & 77.3  \\
\; \alpharelic{} (ours) & {\bf 79.3}  \\  
\bottomrule
\end{tabular}
\subcaption{ResNet152 encoder.}
\end{minipage}
\begin{minipage}{0.5\linewidth}
\centering
\begin{tabular}{lc}
\toprule
Method & Top-1  \\
\midrule
\; Supervised \tiny\citep{grill2020bootstrap} & 79.3  \\ 
\; BYOL \tiny\citep{grill2020bootstrap} & 77.8  \\
\; \alpharelic{} (ours) & {\bf 79.8}  \\
\bottomrule
\end{tabular}
\subcaption{ResNet200 encoder.}
\end{minipage}%
\begin{minipage}{0.5\linewidth}
\centering
\begin{tabular}{lc}
\toprule
Method & Top-1 \\
\midrule
\; Supervised \tiny\citep{grill2020bootstrap} &  80.1  \\ 
\; BYOL \tiny\citep{grill2020bootstrap} &  79.6  \\
\; \alpharelic{} (ours) & {\bf 80.6}  \\
\bottomrule
\end{tabular}
\subcaption{ResNet200 $2\times$ encoder.}
\end{minipage}
\caption{Top-1 accuracy (in \%) under linear evaluation on the ImageNet test set for a varied set of ResNet architectures.}
\label{tab:linear_other_resnets}
\end{table}

\newpage

\begin{wrapfigure}{r}{0.6\textwidth}
    \centering
    \includegraphics[width=0.5\textwidth]{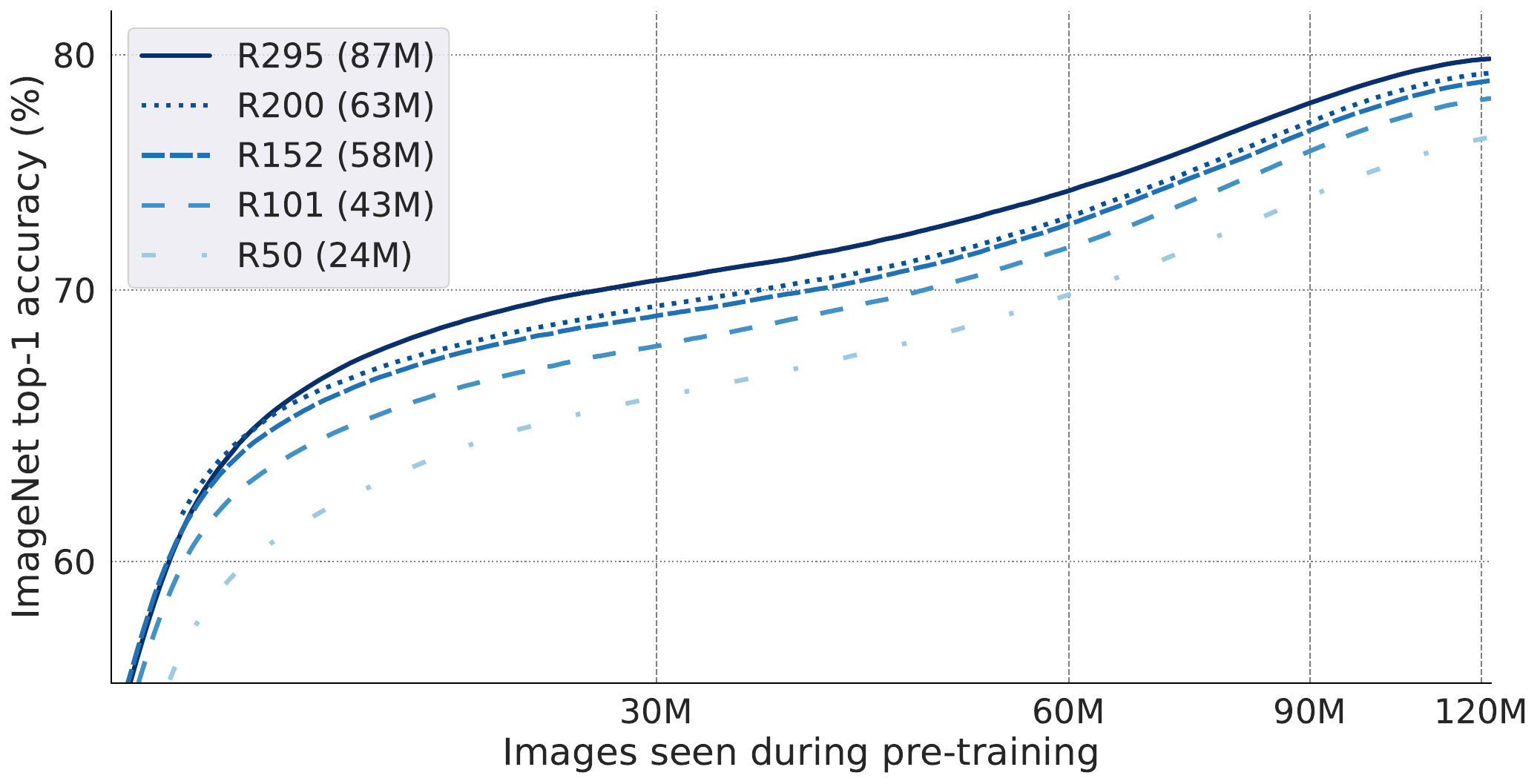}
  \caption{ImageNet accuracy obtained by \alpharelic{} as a function of number of images seen during pre-training for several of ResNet architectures (Number of model parameters in brackets).
  \vspace{-0.2cm}
  \label{fig:sup:scaling}}
\end{wrapfigure}

\noindent\textbf{Scaling.} 
Figure~\ref{fig:sup:scaling} shows the ImageNet linear evaluation accuracy obtained by \alpharelic{} representations as a function of the number of images seen during pre-training.
We see that in order to reach $70\%$ accuracy the ResNet50 model requires approximately twice as many iterations as the ResNet295 model, a model we constructed so that it has approximately $3.6\times$ the number of parameters as the ResNet50 (87M vs 24M, respectively). 
This finding is in accordance with other works which show that larger models are more sample efficient (i.e. they require fewer samples to reach a given accuracy) \citep{zhai2021scaling}.

\subsection{Analysis}
\label{sec:analysis}

\begin{wrapfigure}{r}{0.45\textwidth}
\vspace{-0.4cm}
    \centering
\includegraphics[width=0.4\textwidth]{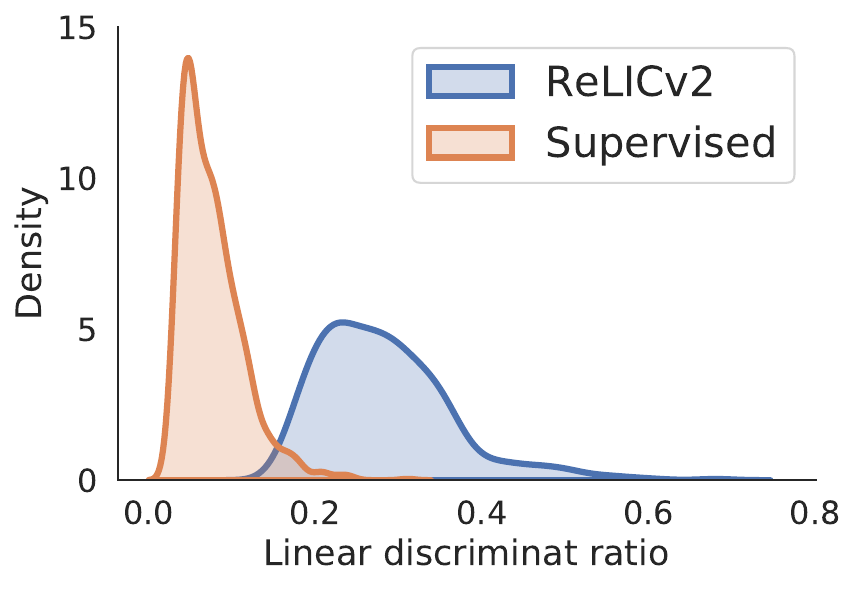}
  \caption{Distribution of the \emph{linear discriminant ratio}: the ratio of between-class distances and within-class distances of embeddings computed on the ImageNet test set. }
\label{fig:relic_kde}
\vspace{-0.3cm}
\end{wrapfigure}

\noindent\textbf{Class concentration.} To quantify the overall structure of the learned latent space, we examine within- and between-class distances. Figure~\ref{fig:relic_kde} compares the distribution of ratios of between-class and within-class $\ell_2$-distances of the \alpharelic{} representations on the ImageNet test set against those learned by the supervised 
 baseline.\footnote{Both \alpharelic{} and the supervised baseline are trained on the ImageNet training set.}  
A higher mean of this distribution means that the representations are better concentrated within the corresponding classes as well as better separated between classes and therefore more easily linearly separable (c.f.~Fisher's linear discriminants \citep{friedman2001elements}). 
We see that \alpharelic's distribution is shifted to the right, i.e. has a higher mean compared to the supervised baseline, suggesting that better representations have been learned. For additional analysis of the  distances between  learned  representations  of closely  related  classes, refer to the appendix.

\noindent\textbf{Views of varying sizes.} Most prior work uses 2 views of size $224 \times 224$ to learn representations.
RELICv2 proposes instead the use of a large number of views of size $224 \times 224$ combined with a
small number of smaller views of size $96 \times 96$. We ablate the use of different numbers of large and
small views in RELICv2 using only SimCLR augmentations (i.e. without saliency masking). We report below the top-1 ImageNet test performance (under linear evaluation) of a ResNet50 pretrained
for 1000 epoch on ImageNet; $[L, S]$ denotes using L large views and S small views.

\begin{center}
\begin{adjustbox}{max width=\columnwidth}
\begin{tabular}{l|cccccccc}
Views & [2, 0] & [2, 2] & [2, 6] & [4, 0] & [4, 2] & [6, 2] & [8, 2] \\
\hline
Top-1 & 74.8 & 76.2 & 76.0 & 75.5 & {\bf 76.8} & 76.5 & 76.5 \\
\end{tabular}
\end{adjustbox}
\end{center}

There seems to be a performance plateau going beyond 6 large views and a slight performance
penalty going beyond 4 large views. For small views, we also observe performance penalties going
beyond 2 small views, while there is a significant performance boost going from no small views to
2 small views, i.e. $+1.3\%$ in the case of 4 large views. This is almost double the performance improvement one gets from adding 2 large views, i.e the difference between $[2, 0]$ and $[4, 0]$ of $+0.7\%$. This finding supports our hypothesis that small views significantly contribute to learning better representations, but that we only need a small number of them. Note
that our use of large and small views is exactly the opposite as compared to SWaV \citep{caron2018deep} which argue for using smaller views as computationally less expensive alternatives to large
views and rely heavily on using a large number of small views; in particular, they argue for using 2
large views and 6 small views.

\noindent{\textbf{Saliency masking}}. To isolate the contribution of saliency masking, we measure the performance
gain (in terms of top-1 accuracy under linear evaluation on ImageNet) when applying saliency masking to just 2 large views. This improves performance from $74.8\%$ to $75.3\%$, i.e. a gain of $+0.5\%$
which is a boost comparable to having two additional large views (see above). We also explored
using different datasets for pretraining our unsupervised saliency masking pipeline as well as varying the probability with which we apply saliency masking to the inputs and found that RELICv2 is
robust to these choices; see the appendix for details.

\begin{wraptable}{r}{0.5\textwidth}
\vspace{-0.5cm}
\begin{center}
\begin{adjustbox}{max width=0.5\textwidth}
\begin{tabular}{l|l}
 & Top-1 \\
\hline
\; SimCLR {\tiny\citep{chen2020simple}}  & 64.5\% \\
\; SimCLR + [4, 2] + saliency & 66.2\%  (+1.7\%) \\
\hline
\; \relic{} {\tiny\citep{mitrovic2020representation}} & 61.1\%  \\
\; \alpharelic{} (ours) & 67.5\% (+6.4\%) \\
\hline
\end{tabular}
\end{adjustbox}
\end{center}
\vspace{-0.4cm}
\end{wraptable}

\noindent\textbf{Invariance}. To assess the importance of enforcing invariance over background removal and object
styles, we extend SimCLR to use saliency masking, 4 large views and 2 small views, and compare it
to RELICv2. 
We train both methods for 100 epochs
and report top-1 accuracy on ImageNet. 
As we can see from the table, invariance-based methods
profit from multiple views and saliency masking significantly more than purely contrastive methods.
Thus, invariance plays a crucial role in learning better representations.

\vspace{-0.2cm}
\section{Additional experimental results}
\vspace{-0.2cm}
\label{sec:res}

We showcase the generality and wide applicability of \alpharelic{} on transfer, semi-supervised, robustness and out-of-distribution (OOD) generalization tasks when pretraining on ImageNet as well as when we pretrain on the much larger and more complex Joint Foto Tree (JFT-300M) dataset \citep{ sun2017revisiting}. 
Furthermore, we test the transferability of \alpharelic{} representations to more complex vision tasks such as semantic segmentation.
For a complete set of results and experimental details please refer to the appendix.

\subsection{Robustness and OOD generalization}
\label{sec:res:robust}

We evaluate the robustness and out-of-distribution (OOD) generalization of \alpharelic{} on a wide variety of datasets.
We use ImageNetV2 \citep{recht2019imagenet} and ImageNet-C \citep{hendrycks2019benchmarking} to evaluate robustness, while ImageNet-R \citep{hendrycks2021many}, ImageNet-Sketch \citep{wang2019learning} and ObjectNet \citep{barbu2019objectnet} are used to evaluate OOD generalization. 
We evaluate \alpharelic{} representations pretrained on ImageNet on ResNet50 under linear protocol akin to Section \ref{sec:results:linear}, i.e.\ we train a linear classifier on top of the frozen representation using the labelled ImageNet training set; the test evaluation is performed zero-shot, i.e no training is done on the above datasets. 
From Table \ref{table.imagenet_v2_c}, we see that \alpharelic{}
outperforms competing self-supervised models by a wide margin in terms of robustness on ImageNetV2 and ImageNet-C. 
Furthermore, \alpharelic{} also shows better robustness performance than the supervised baseline.
In terms of OOD generalization, \alpharelic{} also learns representations that outperform competing self-supervised methods while being on par with supervised performance. 
For a detailed explanation of the datasets and a full breakdown of the results please see the appendix.

\begin{table}[h]
 \begin{center}
 \vspace{-0.2cm}
\caption{Top-1 Accuracy (in \%) under linear evaluation on ImageNetV2 and ImageNet-C (robustness), and ImageNet-R (IN-R), ImageNet-Sketch (IN-S) and ObjectNet (out-of-distribution). ImageNetv2 has three variants -- matched frequency (MF), Threshold 0.7 (T-0.7) and Top Images (TI). The results for ImageNet-C (IN-C) are averaged across the 15 different corruptions.} \label{table.imagenet_v2_c}
 \vspace{-0.1cm}
 \begin{tabular}{lcccc|ccc}
 \toprule
 & \multicolumn{4}{c}{Robustness} & \multicolumn{3}{c}{OOD Generalization} \\
 \toprule
 Method & MF & T-0.7 & Ti & IN-C  & IN-R & IN-S & ObjectNet   \\
 \toprule
 \; Supervised  & 65.1  & 73.9 & 78.4 & 40.9 & 24.0 & 6.1  & 26.6 \\ 
 \midrule
 \; SimCLR \citep{chen2020simple}  & 53.2 & 61.7 & 68.0 & 31.1 & 18.3 & 3.9 & 14.6 \\
 \; BYOL \citep{grill2020bootstrap} & 62.2 & 71.6  & 77.0 & 42.8   & 23.0 & 8.0  &  23.0  \\
 \; \relic{} \citep{mitrovic2020representation} & 63.1 & 72.3 &  77.7 & 44.5 & 23.8 & 9.1 & 23.8 \\
 \; \alpharelic{} (ours) & \textbf{65.3} & \textbf{74.5} & \textbf{79.4} & \textbf{44.8} & \textbf{23.9} & \textbf{9.9} & \textbf{25.9}  \\ 
 \bottomrule
 \end{tabular}
 \end{center}
 
 \vspace{-0.6cm}
\end{table}

\subsection{Transfer to other classification datasets and semantic segmentation}
\label{sec:res:transfer}

\begin{figure}[h]
\vspace{-0.1cm}
    \centering
    \includegraphics[width=0.9\textwidth]{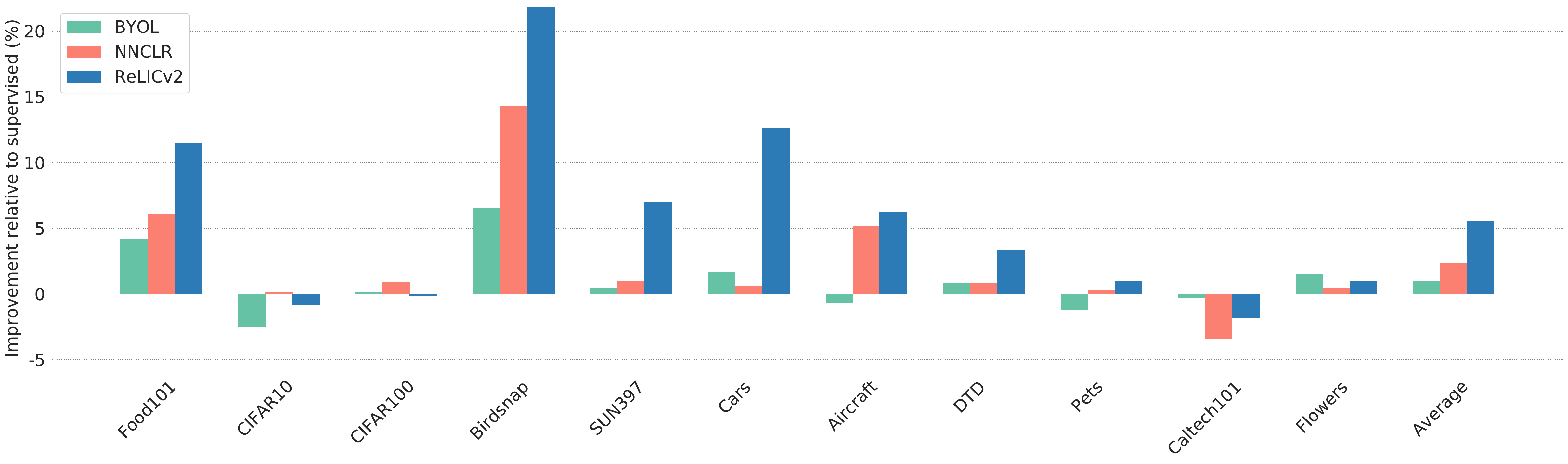}
  \caption{Transfer performance relative to the supervised baseline (a value of 0 indicates equal performance to supervised).}
\label{fig:transfer}
\vspace{-0.2cm}
\end{figure}

\paragraph{Classification.} We perform linear evaluation and fine-tuning on a wide range of classification benchmarks. 
We follow established evaluation protocols from the literature as described in the appendix and report standard metrics for each dataset on held-out test sets. 
Figure~\ref{fig:transfer} compares the transfer performance of BYOL, NNCLR and \alpharelic{} pre-trained representations relative to supervised pre-training. 
\alpharelic{} improves upon both the supervised baseline and competing methods, performing best on 7 out of 11 tasks. \alpharelic{} has an average relative improvement
of over $+5\%$ across all tasks---over double that of closest competing self-supervised method NNCLR. 
For detailed results on linear evaluation and fine-tuning see appendix. 

\paragraph{Semantic Segmentation.}

\begin{wraptable}{r}{0.5\textwidth}
\vspace{-0.6cm}
\begin{center}
\caption{Transfer performance on semantic segmentation tasks.}
\vspace{-0.3cm}
\label{table.transfer_other_main}
\begin{adjustbox}{max width=0.5\textwidth} 
\centering
\begin{tabular}{l c c}
\toprule
Method & PASCAL & Cityscapes \\
\midrule
BYOL \citep{grill2020bootstrap}  & 75.7 & 74.6 \\
ReLICv2 (ours) & {\bf 77.9} & {\bf 75.2} \\
\bottomrule
\end{tabular}
\end{adjustbox}
\vspace{-0.3cm}
\end{center}
\end{wraptable}

We evaluate the ability of \alpharelic{} to facilitate successful transfer of the learned representations to the PASCAL \citep{Everingham10} and Cityscapes \citep{Cordts2016Cityscapes} semantic segmentation tasks. 
In accordance with~\citep{he2019momentum}, we initialize a fully convolutional backbone with the \alpharelic{} model pretrained on ImageNet.
We fine-tune that model on the PASCAL \texttt{train\_aug2012} set for 45 epochs and report the mean intersection over union (mIoU) on the \texttt{val2012} set. 

The fine-tuning on Cityscapes is done on the \texttt{train\_fine} set for 160 epochs and evaluated on the \texttt{val\_fine} set. 
From Table \ref{table.transfer_other_main} we see that \alpharelic{} significantly outperforms BYOL on both datasets. 
Interestingly, \alpharelic{} outperforms even DetCon \citep{detcon}, a method specifically trained for detection, on PASCAL where it achieves $77.3$ (mIoU).

\subsection{Pretraining on Joint Foto Tree (JFT-300M)} \label{app.results_jft}

We also test how well \alpharelic{} scales to much larger datasets by pretraining representations using the Joint Foto Tree (JFT-300M) dataset which consists of 300 million images from more than 18k classes \citep{ sun2017revisiting}.
We  evaluate the learned representations on the ImageNet test set under the same linear evaluation protocol as described in Section \ref{sec:results:linear}. 

\begin{wraptable}{r}{0.7\textwidth}
\vspace{-0.3cm}
\caption{Top-1 accuracy (in \%) on ImageNet when learning representations using the JFT-300M dataset.}
\vspace{-0.2cm}
\label{table.jft-top1_new}
\begin{center}
\begin{adjustbox}{max width=0.7\columnwidth}
\begin{tabular}{lcccccc}
\toprule
& \multicolumn{3}{c}{Top-1} \\
Method & 1000 epochs  & 3000 epochs & $>$4000 epochs \\
\midrule
\; BYOL \tiny\citep{grill2020bootstrap}  &  67.0 & 67.6 & 67.9 (5000 epochs)\\
\; Divide and Contrast \tiny\citep{tian2021divide} & 67.9 & 69.8  & 70.7 (4500 epochs) \\
\; \alpharelic{} (ours) & \textbf{70.3}  & \textbf{71.1} & \textbf{71.4} (5000 epochs) \\ 
\bottomrule
\end{tabular}
\end{adjustbox}
\end{center}
\vspace{-0.5cm}
\end{wraptable}

We compare RELICv2 against BYOL and Divide and Contrast \citep{tian2021divide}, a method
that was specifically designed to handle large and uncurated datasets and represents the current
state-of-art in self-supervised JFT-300M pretraining. Table \ref{table.jft-top1_new} reports the top-1 accuracy when training the various methods using ResNet50 as the backbone for different number of ImageNet equivalent epochs on JFT-300M.
\alpharelic{} improves over Divide and Contrast \citep{tian2021divide} by more than $+2\%$ when training on JFT for 1000 epochs and achieves better overall performance than competing methods while needing a smaller number of training epochs. 
Please refer to the appendix for experimental details and additional results on JFT-300M such as transfer learning, robustness and out-of-distribution evaluation.

\section{Related work} \label{sec:review}

Recently, \emph{contrastive} approaches have become an important area of research owing to their excellent performance in visual recognition tasks \citep{oord2018representation, bachman2019learning, chen2020simple, he2019momentum, dwibedi2021little, grill2020bootstrap}. 
Moreover, bootstrapping-based learning has also achieved comparable performance \citep{grill2020bootstrap}.
The aforementioned methods implicitly enforce invariance by maximizing the similarity between two views. 
On the other hand, \citep{caron2020unsupervised} incorporates an explicit clustering step as a more direct way of enforcing some notion of invariance.
However, neither of these strategies can be directly linked theoretically to learning more compact representations.  
\citep{mitrovic2020representation} approach invariance from a causal perspective and use an explicit invariance loss in conjunction with a contrastive objective to learn representations.
They show that invariance must be explicitly enforced---via an invariance loss in addition to a contrastive loss---in order to obtain guaranteed generalization performance. 

Background augmentations have recently been proposed as part of the data augmentation pipeline in \citep{zhao2020distilling} and further discussed in concurrent work to our manuscript \citep{ryali2021leveraging}. 
\alpharelic{} differs from these methods in two key points. 
First, we enforce an explicit invariance loss between views with and without background, while \citep{zhao2020distilling} do not have any invariance loss.
Second, \alpharelic{} learns representations using multiple views of varying sizes, while these methods use only two views. 
As a computationally favourable alternative to large views, \citep{caron2020unsupervised} propose to use a large number of small views in addition to the two large views. 
In a notable difference to \citep{caron2020unsupervised}, \alpharelic{} employs small views to increase the representation robustness and proposes to use only a small number of small views. 
Additionally, \alpharelic{} imposes an invariance loss between pairs of large and small views, and uses a large number of large views.
Outside of contrastive learning, \citep{lee2021compressive} take a conditional entropy bottleneck approach which results in better-than-supervised performance albeit only on the ResNet50 (2x) architecture (see Figure~\ref{fig:top1}).
Please refer to the appendix for a more detailed comparison of related works.

\section{Discussion}
\label{sec:discussion}

\begin{wrapfigure}{r}{0.45\textwidth}
\vspace{-0.3cm}
    \centering
    \includegraphics[width=0.45\textwidth]{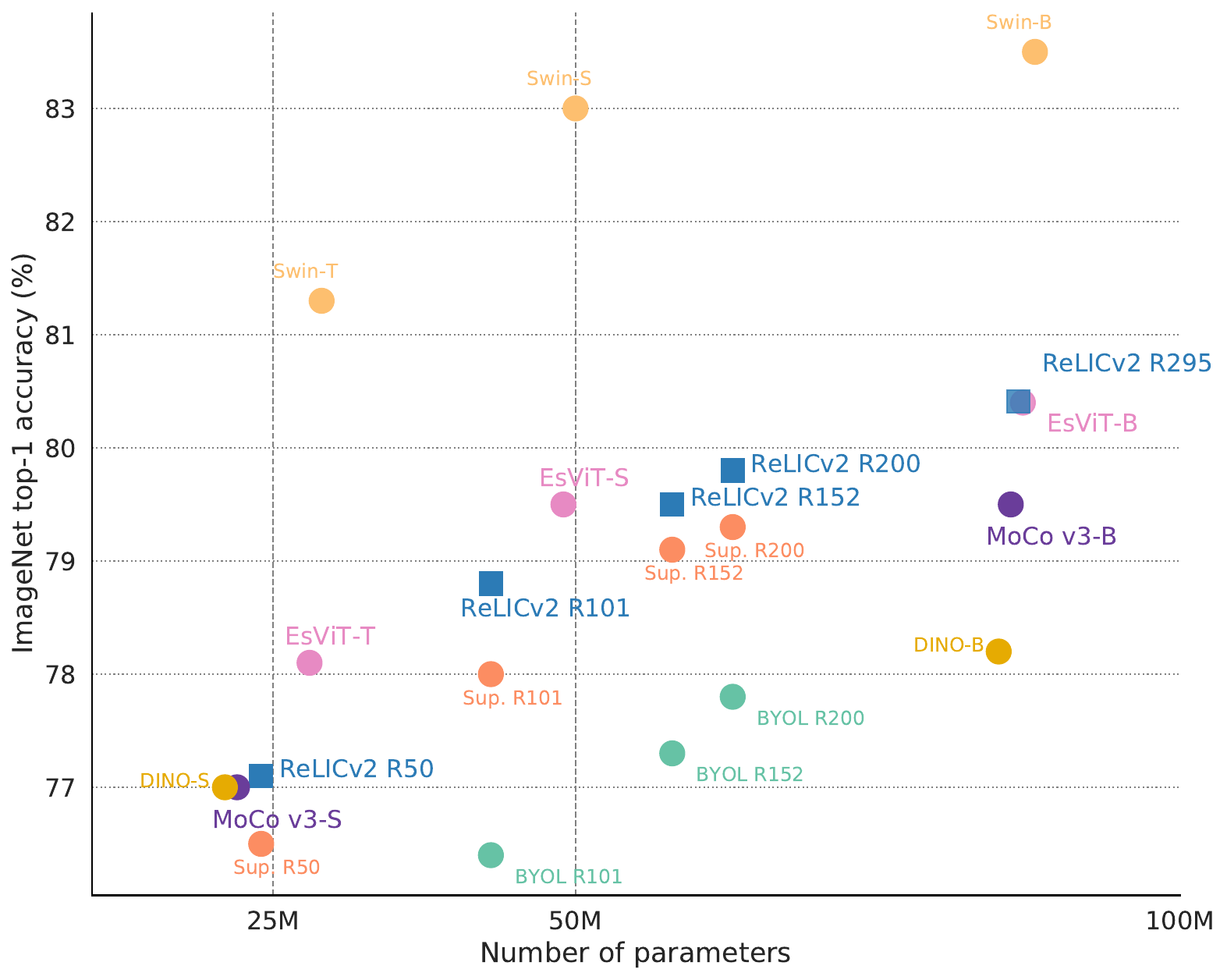}
  \caption{Comparison of ImageNet top-1 accuracy between \alpharelic{} and recent vision transformer-based architectures (Swin \citep{liu2021swin} is a fully supervised transformer baseline).
  \label{fig:top1_vit}}
  \vspace{-0.3cm}
\end{wrapfigure}

We proposed a novel self-supervised representation learning method \alpharelic{}, which learns representations by enforcing invariance over views of various sizes and uses saliency masking as part of the data augmentations pipeline to learn representations that are robust to spurious correlations. 
\alpharelic{} sets the new state-of-the-art in self-supervised learning by significantly outperforming previous methods across a wide range of ResNet architectures by up to $+2.3\%$ in linear evaluation on ImageNet. 
Moreover, \alpharelic{} achieves state-of-the art performance on transfer and semi-supervised learning and robustness and out-of-distribution generalization tasks. 
We also show that \alpharelic{} sets a new state-of-the-art when pretrained on JFT-300M and in the process outperforms competing self-supervised methods specifically developed for large and uncurated datasets such as JFT-300M.
On semantic segmentation, our method outperforms competing self-supervised methods, some of them specifically designed for object detection, unlike \alpharelic{}. 
Furthermore, \alpharelic{} is the first method that demonstrates that representations learned without access to labels can consistently outperform a standard supervised baseline on ImageNet which is a first step in making self-supervised learning the go-to approach to learning performant representations.


Vision transformers (ViTs) \citep{dosovitskiy2020image} have recently emerged as promising architectures for visual representation learning. 
Figure~\ref{fig:top1_vit} compares recent ViT-based methods against \alpharelic{} using a variety of larger ResNet architectures. 
Notably, \alpharelic{} outperforms recent self-supervised ViT-based methods DINO \citep{caron2021emerging} and MoCov3 \citep{chen2021empirical} as well as exhibits similar performance to EsViT \citep{li2021efficient} for comparable parameter counts despite the fact that these methods use more powerful architectures and more involved training procedures. 
Our results suggest that combining the insights we have developed with \alpharelic{} alongside recent architectural innovations could lead to further improvements in representation learning and more powerful foundation models.

\subsubsection*{Acknowledgments}
We thank Relja Arandjelovic, Yusuf Aytar, Andrew Zisserman, Koray Kavukcuoglu, Daan Wierstra and Karen
Simonyan for discussions on this work and feedback on the
manuscript.

\bibliographystyle{plainnat}
\bibliography{relic}

\appendix

\newpage
\section{Comparison between self-supervised methods}
\label{sec:sup_review}

In this review we focus on how important algorithmic choices: namely explicitly enforcing invariance and more considered treatment of positive and negative examples are key factors in improving downstream classification performance of unsupervised representations.

\begin{table}[h]
\begin{center}
\begin{adjustbox}{max width=\columnwidth}
\begin{tabular}{lccll}
\toprule
Method & Contrastive & Invariance & Positives & Negatives \\
\midrule
SimCLR \citep{chen2020simple}          & \Checkmark & \XSolid & $\text{aug}(x_i)$     & full batch            \\ [0.25cm]
BYOL \citep{grill2020bootstrap}           & \XSolid & $\ell_2$ &  $\text{aug}(x_i)$    & n/a           \\ [0.25cm]
NNCLR \citep{dwibedi2021little}           & \Checkmark & \XSolid & $\text{aug}(x_i)$, $\text{nn}(x_i)$  & full batch \\ [0.4cm]
MoCo \citep{he2019momentum}  & \Checkmark & \XSolid & $\text{aug}(x_i)$  & queue \\ [0.25cm]
SwAV \citep{caron2020unsupervised}   & \Checkmark & \XSolid & $\text{aug}(x_i)$, $\text{mc}(x_i)$, $\text{proto}^{+}(x_i)$  & $\text{proto}^{-}(x_i)$ \\ [0.25cm]
Debiased \citep{chuang2020debiased}  & \Checkmark & \XSolid & $\text{aug}(x_i)$  & importance sample \\ [0.4cm]
Hard Negatives \citep{robinson2020contrastive}         & \Checkmark & \XSolid & $\text{aug}(x_i)$  & importance sample \\ [0.4cm]
ReLICv1 \citep{mitrovic2020representation}   & \Checkmark & ${D}_\text{KL}$ & $\text{aug}(x_i)$  & subsample \\ [0.25cm]
\alpharelic{} (ours)  & \Checkmark & ${D}_\text{KL}$ & $\text{aug}(x_i)$, $\text{mc}(x_i)$, $\text{sal}(x_i)$ & subsample \\ [0.25cm]
\bottomrule
\end{tabular}
\end{adjustbox}
\end{center}
\caption{The role of positives and negatives in recent unsupervised representation learning algorithms.}
\label{table:methods}
\end{table}

\paragraph{Negatives.}
A key observation of \citep{chen2020simple} was that large batches (up to 4096) improve results. This was partly attributed to the effect of more negatives. This motivated the incorporation of queues that function as large reservoirs of negative examples into contrastive learning \citep{he2019momentum}.
However subsequent work has shown that naively using a large number of negatives can have a detrimental effect on learning \citep{mitrovic2020less, saunshi2019theoretical, chuang2020debiased, robinson2020contrastive}. One reason for this is due to \emph{false negatives}, that is points in the set of negatives which actually belong to the same latent class as the anchor point. These points are likely to have a high relative similarity to the anchor under $\phi$ and therefore contribute disproportionately to the loss. This will have the effect of pushing apart points belonging to the same class in representation space.  The selection of true negatives is a difficult problem as in the absence of labels it necessitates having access to reasonably good representations to begin with. As we do not have access to these representations, but are instead trying to learn them, there has been limited success in avoiding false and selecting informative negatives.
This phenomenon explains the limited success of attempts to perform hard negative sampling. 

Subsampling-based approaches have been proposed to avoid false negatives via importance sampling to attempt to find \emph{true} negatives which are close to the latent class boundary of the anchor point \citep{robinson2020contrastive}, or uniformly-at-random sampling a small number of points to avoid false negatives \citep{mitrovic2020less}. 

\paragraph{Positives and invariance.}
Learning representations which are invariant to data augmentation is known to be important for self-supervised learning. Invariance is achieved heuristically through comparing two different augmentations of the same anchor point. Incorporating an explicit clustering step is another way of enforcing some notion of invariance \citep{caron2020unsupervised}. However, neither of these strategies can be directly linked theoretically to learning more compact representations. More rigorously \citep{mitrovic2020representation} approach invariance from a causal perspective. They show that invariance must be explicitly enforced---via an invariance loss in addition to the contrastive loss---in order to obtain guaranteed generalization performance. 
Most recently \citep{dwibedi2021little} and \citep{assran2021semi} use nearest neighbours to identify other elements from the batch which potentially belong to the same class as the anchor point.

Table \ref{table:methods} provides a detailed comparison in terms of how prominent representation learning methods utilize positive and negative examples and how they incorporate both explicit contrastive and invariance losses. Here $\text{aug}(x_i)$ refers to the standard set of SimCLR augmentations \citep{chen2020simple}, $\text{nn}(x_i)$ refers to a scheme which selects nearest neighbours of $x_i$, $\text{mc}(x_i)$ are multicrop augmentations (c.f.~\citep{caron2020unsupervised}). $\text{proto}^+(x_i)$ and $\text{proto}^-(x_i)$ refer to using prototypes computed via an explicit clustering step c.f.~\citep{caron2020unsupervised}. Finally, $\text{sal}(x_i)$ refers to a scheme which computes saliency masks of $x_i$ and removes backgrounds as described in section~\ref{sec:method}. Note that SwAV first computes a clustering of the batch then contrasts the embedding of the point and its nearest cluster centroid $(\text{proto}^{+})$ against the remaining $K-1$ cluster centroids $(\text{proto}^{-})$; invariance is implicitly enforced in the clustering step.

\newpage
\section{Pretraining on ImageNet -- implementation details and additional results}
\label{sec:supp:add_results}

Similar to previous work \citep{chen2020simple,grill2020bootstrap} we minimize our objective using the LARS optimizer \citep{you2017large} with a cosine decay learning rate schedule without restarts. 
Unless otherwise indicated, we train our models for $1000$ epochs with a warm-up period of $10$ epochs and a batch size of $|\mathcal{B}|=4096$. 
In our experiments, we use $4$ views of the standard size $224\times 224$ and $2$ views of the smaller size $96\times96$ each coming from an image augmented by a different randomly chosen data augmentation; the smaller size views are centered crops of the randomly augmented image. For a detailed ablation analysis on the number of large and small crops see section \ref{sec:analysis}.

\subsection{Linear evaluation}  \label{app.linear_eval}

Following the approach of \citep{chen2020simple,grill2020bootstrap,caron2020unsupervised,dwibedi2021little}, we use the standard linear evaluation protocol on ImageNet. We train a linear classifier on top of the frozen representation which has been pretrained, i.e. the encoder parameters as well as the batch statistics are not being updated. For training the linear layer, we preprocess the data by applying standard spatial augmentations, i.e. randomly cropping the image with subsequent resizing to $224\times 224$ and then randomly applying a horizontal flip. At test time, we resize images to 256 pixels along the shorter side with bicubic resampling and apply a $224\times 224$ center crop to it. Both for training and testing, after performing the above processing, we normalize the color channels by substracting the average channel value and dividing by the standard deviation of the channel value (as computed on ImageNet). To train the linear classifier, we optimize the cross-entropy loss with stochastic gradient descent with Nestorov momentum for $100$ epochs using a batch size of $1024$ and a momentum of $0.9$; we do not use any weight decay or other regularization techniques. 

\begin{wraptable}{r}{0.45\textwidth}
\vspace{-0.5cm}
\begin{center}
\begin{adjustbox}{max width=0.45\textwidth}
\begin{tabular}{lccc}
\toprule
Method & Top-1 & Top-5 \\
\toprule
\; Supervised \tiny\citep{chen2020simple}   & 76.5 & 93.7 \\ 
\midrule
\; SimCLR {\tiny\citep{chen2020simple}}  & 69.3 & 89.0 \\
\; MoCo v2 \tiny\citep{Chen2020ImprovedBW} & 71.1 &  - \\
\; InfoMin Aug. \tiny\citep{Tian2020WhatMF} & 73.0 &  91.1 \\
\; BYOL \tiny\citep{grill2020bootstrap} & 74.3 & 91.6 \\
\; \relic{} \tiny\citep{mitrovic2020representation} & 74.8 & 92.2 \\ 
\; SwAV \tiny\citep{caron2020unsupervised} & 75.3 & - \\
\; NNCLR  \tiny\citep{dwibedi2021little}  & 75.6 & 92.4 \\
\; C-BYOL \tiny\citep{lee2021compressive} & 75.6 & 92.7 \\
\; \alpharelic{} (ours) & \textbf{77.1} & 93.3 \\
\hline
\end{tabular}
\end{adjustbox}
\end{center}
\caption{Top-1 and top-5 accuracy (in \%) under linear evaluation on the ImageNet test set for a ResNet50 encoder set for different representation learning methods.}
\label{table.imagenet_linear_all}
\vspace{-0.5cm}
\end{wraptable}

In the following tables, we report the top-1 and top-5 accuracies of different methods under a varied set of ResNet encoders of different sizes, spanning ResNet50, ResNet101, ResNet152 and ResNet200 and layer widths of $1\times$, $2\times$ and $4\times$.
ResNet50 with $2\times$ and $4\times$ wider layers has $94$ and $375$ million parameters, respectively.
ResNet101, ResNet152, ResNet200 and ResNet200 $2\times$ have $43$, $58$, $63$ and $250$ million parameters, respectively.

In Table \ref{tab:linear_other}, we present results under linear evaluation on the ImageNet test set a varied set of ResNet architectures; we compare against different unsupervised representation learning methods and use as the supervised baselines the results reported in \citep{chen2020simple,grill2020bootstrap}. Note that the supervised baselines reported in \citep{chen2020simple} are extensively used throughout the self-supervised literature in order to compare performance against supervised learning.
For architectures for which supervised baselines are not available in \citep{chen2020simple}, we use supervised baselines reported in \citep{grill2020bootstrap} which use stronger augmentations for training supervised models than \citep{chen2020simple} and as such do not represent a direct like-for-like comparison with self-supervised methods. Across this varied set of ResNet architectures, \alpharelic{} outperforms all competing self-supervised methods while also outperforming the supervised baselines in all cases with margins up to $1.2\%$ in absolute terms.

\begin{table}[h]
\begin{minipage}{.45\linewidth}
    \centering
\begin{tabular}{lccccc}
\toprule
Method & Top-1 & Top-5 \\
\midrule
\; Supervised \tiny\citep{chen2020simple} &  77.8 & -- \\ 
\; MoCo \tiny\citep{he2019momentum} & 65.4 & -- \\
\; SimCLR \tiny\citep{chen2020simple} & 74.2 & 92.0 \\
\; BYOL \tiny\citep{grill2020bootstrap} &  77.4 & 93.6 \\
\; SwAV \tiny\citep{caron2020unsupervised} &  77.3 & -- \\
\; C-BYOL \tiny\citep{lee2021compressive}  & 78.8 & {\bf 94.5} \\
\; \alpharelic{} (ours) &   {\bf 79.0} & {\bf 94.5} \\
\bottomrule
\end{tabular}
\subcaption{ResNet50 $2\times$ encoder.}
\end{minipage}%
    \begin{minipage}{.65\linewidth}
\centering
\begin{tabular}{lccccc}
\toprule
Method & Top-1 & Top-5 \\
\midrule
\; Supervised \tiny\citep{chen2020simple} &  78.9 & -- \\ 
\; MoCo \tiny\citep{he2019momentum} & 68.6 & -- \\
\; SimCLR \tiny\citep{chen2020simple}  & 76.5 & 93.2 \\
\; SwAV \tiny\citep{caron2020unsupervised} &  77.9 & -- \\
\; BYOL \tiny\citep{grill2020bootstrap} &  78.6 & 94.2 \\
\; \alpharelic{} (ours) &  {\bf 79.4} & {\bf 94.3} \\  
\bottomrule
\end{tabular}
\subcaption{ResNet50 $4\times$ encoder.}
\end{minipage}
\begin{minipage}{0.5\linewidth}
\centering
\begin{tabular}{lccccc}
\toprule
Method & Top-1 & Top-5 \\
\midrule
\; Supervised \tiny\citep{grill2020bootstrap} & 78.0 & 94.0 \\ 
\; BYOL \tiny\citep{grill2020bootstrap} & 76.4 & 93.0 \\
\; \alpharelic{} (ours) & {\bf 78.7} & {\bf 94.4}  \\ 
\bottomrule
\end{tabular}
\subcaption{ResNet101 encoder.}
\end{minipage}%
\begin{minipage}{0.5\linewidth}
\centering
\begin{tabular}{lccccc}
\toprule
Method & Top-1 & Top-5 \\
\midrule
\; Supervised \tiny\citep{grill2020bootstrap} & 79.1 & 94.5 \\ 
\; BYOL \tiny\citep{grill2020bootstrap} & 77.3 & 93.7 \\
\; \alpharelic{} (ours) & {\bf 79.3} & {\bf 94.6}  \\  
\bottomrule
\end{tabular}
\subcaption{ResNet152 encoder.}
\end{minipage}
\begin{minipage}{0.5\linewidth}
\centering
\begin{tabular}{lccccc}
\toprule
Method & Top-1 & Top-5 \\
\midrule
\; Supervised \tiny\citep{grill2020bootstrap} & 79.3 & 94.6  \\ 
\; BYOL \tiny\citep{grill2020bootstrap} & 77.8 & 93.9 \\
\; \alpharelic{} (ours) & {\bf 79.8} & {\bf 95.0}  \\
\bottomrule
\end{tabular}
\subcaption{ResNet200 encoder.}
\end{minipage}%
\begin{minipage}{0.5\linewidth}
\centering
\begin{tabular}{lccccc}
\toprule
Method & Top-1 & Top-5 \\
\midrule
\; Supervised \tiny\citep{grill2020bootstrap} &  80.1 & {\bf 95.2} \\ 
\; BYOL \tiny\citep{grill2020bootstrap} &  79.6 & 94.8 \\
\; \alpharelic{} (ours) & {\bf 80.6} & {\bf 95.2} \\
\bottomrule
\end{tabular}
\subcaption{ResNet200 $2\times$ encoder.}
\end{minipage}
\caption{Top-1 and top-5 accuracy (in \%) under linear evaluation on the ImageNet test set for a varied set of ResNet architectures.}
\label{tab:linear_other}
\end{table}

\newpage
\subsection{Semi-supervised learning}  \label{app.semi_sup_results}
We further test \alpharelic{} representations learned on bigger ResNet models in the semi-supervised setting.
For this, we follow the semi-supervised protocol as in \citep{zhai2019s4l,chen2020simple,grill2020bootstrap,caron2020unsupervised}.
First, we initialize the encoder with the parameters of the pretrained representation and we add on top of this encoder a linear classifier which is randomly initialized. 
Then we train both the encoder and the linear layer using either $1\%$ or $10\%$ of the ImageNet training data; for this we use the splits introduced in \citep{chen2020simple} which have been used in all the methods we compare to \citep{grill2020bootstrap,caron2020unsupervised,dwibedi2021little,lee2021compressive}.
For training, we randomly crop the image and resize it to $224\times 224$ and then randomly apply a horizontal flip. 
At test time, we resize images to 256 pixels along the shorter side with bicubic resampling and apply a $224\times 224$ center crop to it. 
Both for training and testing, after performing the above processing, we normalize the color channels by substracting the average channel value and dividing by the standard deviation of the channel value (as computed on ImageNet). Note that this is the same data preprocessing protocol as in the linear evaluation protocol.
To train the model, we use a cross entropy loss with stochastic gradient descent with Nesterov momentum of $0.9$.
For both $1\%$ and $10\%$ settings, we train for $20$ epochs and decay the initial learning rate by a factor $0.2$ at $12$ and $16$ epochs.
Following the approach of \citep{caron2020unsupervised}, we use the optimizer with different learning rates for the encoder and linear classifier parameters. 
For the $1\%$ setting, we use a batch size of $2048$ and base learning rates of $10$ and $0.04$ for the linear layer and encoder, respectively; we do not use any weight decay or other regularization technique.
For the $10\%$ setting, we use a batch size of $512$ and base learning rates of $0.3$ and $0.004$ for the linear layer and encoder, respectively; we use a weight decay of $1e-5$, but do not use any other regularization technique.

Top-1 and top-5 accuracy on the ImageNet test set is reported in Table \ref{table.imagenet_semi_sup}. 
\alpharelic{} outperforms both the standard supervised baseline and all previous state-of-the-art self-supervised methods when using $10\%$ of the data for fine-tuning. When using $1\%$ of the data, only C-BYOL performs better than \alpharelic{}.
From Table \ref{semi_sup_other}, we see that \alpharelic{} outperforms competing self-supervised methods on ResNet50 $2\times$ in both the $1\%$ and $10\%$ setting. 
For larger ResNets, ResNet50 $4\times$ and ResNet200 $2\times$, \alpharelic{} is state-of-the-art with respect to top-1 accuracy for the low-data regime of $1\%$. 
On these networks for the higher data regime of $10\%$ BYOL outperforms \alpharelic{}.
Note that BYOL trains their semi-supervised models for $30$ or $50$ epochs whereas \alpharelic{} is trained only for $20$ epochs.
We hypothesize that longer training (e.g. $30$ or $50$ epochs as BYOL) is needed for \alpharelic{} representations on larger ResNets as there are more model parameters. 

\begin{table}[t]
\begin{center}
\begin{tabular}{lccccc}
\toprule
Method & \multicolumn{2}{c}{Top-1} & \multicolumn{2}{c}{Top-5} \\
& 1\% & 10\% & 1\% & 10\% \\
\toprule
Supervised \tiny{\citep{zhai2019s4l}}   & 25.4 & 56.4 & 48.4 & 80.4 \\ 
SimCLR \tiny{\citep{chen2020simple}}  & 48.3 & 65.6 & 75.5 & 87.8 \\
BYOL \tiny{\citep{grill2020bootstrap}} & 53.2 & 68.8 & 78.4 & 89.00 \\
SwAV \tiny{\citep{caron2020unsupervised}} & 53.9 & 70.2 & 78.5 & 89.9 \\
NNCLR \tiny{\citep{dwibedi2021little}}  & 56.4 & 69.8 & 80.7 & 89.3 \\
C-BYOL \tiny{\citep{lee2021compressive}} & {\bf 60.6} & 70.5 & {\bf  83.4}& 90.0 \\
\alpharelic{} (ours) & 58.1 & {\bf 72.4} & 81.3 & {\bf 91.2} \\   
\hline
\end{tabular}
\end{center}
\caption{Top-1 and top-5 accuracy (in \%) on the ImageNet test set after semi-supervised training with a fraction of ImageNet labels on a ResNet50 encoder for different representation learning methods.}
\label{table.imagenet_semi_sup}
\end{table}

\begin{table}[!h]
\centering
\begin{tabular}{lcccccc}
\toprule
Method & \multicolumn{2}{c}{Top-1} & \multicolumn{2}{c}{Top-5} \\
& 1\% & 10\% & 1\% & 10\% \\
\midrule
\emph{ResNet50 $2\times$ encoder} \\
   \quad SimCLR \tiny{\citep{chen2020simple}}  & 58.5 & 71.7 &83.0 & 91.2 \\
   \quad BYOL \tiny{\citep{grill2020bootstrap}} & 62.2 & 73.5 & 84.1 & 91.7 \\
   \quad \alpharelic{} (ours)  & {\bf 64.7} & {\bf 73.7} & {\bf 85.4} & {\bf 92.0} \\ 
\midrule
\emph{ResNet50 $4\times$ encoder} \\
   \quad SimCLR \tiny{\citep{chen2020simple}} & 63.0 & 74.4 & 85.8 & 92.6 \\
   \quad BYOL \tiny{\citep{grill2020bootstrap}} & 69.1 & {\bf 75.7} & {\bf 87.9} & {\bf 92.5} \\
   \quad \alpharelic{} (ours) & {\bf 69.5} & 74.6 & 87.3 & 91.6 \\ 
\midrule
\emph{ResNet200 $2\times$ encoder} \\
    \quad BYOL \citep{grill2020bootstrap}&  71.2 & {\bf 77.7} & {\bf 89.5} & {\bf 93.7} \\
    \quad \alpharelic{} (ours) & {\bf 72.1} &  76.4 & {\bf 89.5} & 93.0 \\
\bottomrule
\end{tabular}
\caption{Top-1 and top-5 accuracy (in \%) after semi-supervised training with a fraction of ImageNet labels for different ResNet encoders and  unsupervised representation learning methods. Results are reported on the ImageNet test set.}
\label{semi_sup_other}
\end{table}

\subsection{Transfer}
\label{sec:supp:transfer}
We follow the transfer performance evaluation protocol as outlined in \citep{grill2020bootstrap,chen2020simple}.
We evaluate \alpharelic{} in both transfer settings -- linear evaluation and fine-tuning.
For the linear evaluation protocol we freeze the encoder and train only a randomly initialized linear classifier which is put on top of the encoder.
On the other hand, for fine-tuning in addition to training the randomly initialized linear classifier, we also allow for gradients to propagate to the encoder which has been initialized with the parameters of the pretrained representation. 
In line with prior work \citep{chen2020simple,grill2020bootstrap,dwibedi2021little}, we test \alpharelic{} representations on the following datasets: Food101~\citep{bossard2014food}, CIFAR10~\citep{krizhevsky2009learning}, CIFAR100~\citep{krizhevsky2009learning}, Birdsnap~\citep{berg2014birdsnap}, SUN397 (split 1)~\citep{xiao2010sun}, DTD (split 1)~\citep{cimpoi2014describing}, Cars~\citep{krause20133d} Aircraft~\citep{maji2013fine}, Pets~\citep{parkhi2012cats}, Caltech101~\citep{fei2004learning}, and Flowers~\citep{nilsback2008automated}.

Again in line with previous methods \citep{chen2020simple,grill2020bootstrap,dwibedi2021little}, for Food101~\citep{bossard2014food}, CIFAR10~\citep{krizhevsky2009learning}, CIFAR100~\citep{krizhevsky2009learning}, Birdsnap~\citep{berg2014birdsnap}, SUN397 (split 1)~\citep{xiao2010sun}, DTD (split 1)~\citep{cimpoi2014describing}, and Cars~\citep{krause20133d} we report the Top-1 accuracy on the test set, and for Aircraft~\citep{maji2013fine}, Pets~\citep{parkhi2012cats}, Caltech101~\citep{fei2004learning}, and Flowers~\citep{nilsback2008automated} we report the mean per-class accuracy as the relevant metric in the comparisons. For DTD and SUN397, we only use the first split, of the 10 provided splits in the dataset as per \citep{chen2020simple,grill2020bootstrap,dwibedi2021little}.

We train on the training sets of the individual datasets and sweep over different values of the models hyperparameters. 
To select the best hyperparameters, we use the validation sets of the individual datasets. 
Using the chosen hyperparameters, we train the appropriate using the merged training and validation data and test on the held out test data in order to obtain the numbers reported in Table \ref{table.transfer}.
We swept over learning rates \{$.01$, $0.1$, $0.2$, $0.25$, $0.3$, $0.35$, $0.4$, $1.$, $2.$\}, batch sizes \{$128$, $256$, $512$, $1024$\}, weight decay between \{$1\mathrm{e}{-6}$, $1\mathrm{e}{-5}$, $1\mathrm{e}{-4}$, $1\mathrm{e}{-3}$, $0.01$, $0.1$\}, warmup epochs \{$0$, $10$\}, momentum \{$0.9$, $0.99$\}, Nesterov \{True, False\}, and the number of training epochs. For linear transfer we considered setting epochs among \{$20$, $30$, $60$, $80$, $100$\}, and for fine-tuning, we also considered \{$150$, $200$, $250$\}, for datasets where lower learning rates were preferable. 
Models were trained with the SGD optimizer with momentum.

As can be seen from Table \ref{table.transfer}, \alpharelic{} representations yield better performance than both state-of-the-art self-supervised method as well as the supervised baseline across a wide range of datasets. 
Specifically, \alpharelic{} is best on 7 out of 11 datasets and on 8 out of 11 datasets in the linear and fine-tuning settings, respectively.

\begin{table}[h]
\begin{center}
\setlength\tabcolsep{2pt}
\begin{adjustbox}{max width=\columnwidth}
\begin{tabular}{lccccccccccc}
\midrule
Method & Food101 & CIFAR10 & CIFAR100 & Birdsnap & SUN397  & Cars & Aircraft & DTD & Pets & Caltech101 & Flowers \\
\midrule
\multicolumn{12}{l}{\emph{Linear evaluation:}}\\
\midrule
Supervised-IN \tiny{\citep{chen2020simple}} &$72.3$&$93.6$&$78.3$&$53.7$&$61.9$& $66.7$&$61.0$&$74.9$&$91.5$&\textbf{94.5}&$94.7$ \\
SimCLR \tiny{\citep{chen2020simple}}  &$68.4$&$90.6$&$71.6$&$37.4$&$58.8$&$50.3$ &$50.3$&$74.5$&$83.6$&$90.3$&$91.2$\\
BYOL \tiny{\citep{grill2020bootstrap}} &$75.3$&$91.3$&$78.4$&$57.2$&$62.2$&$67.8$&$60.6$&$75.5$&$90.4$&$94.2$& \textbf{96.1}\\
NNCLR \tiny{\citep{dwibedi2021little}} &$76.7$&\textbf{93.7}&\textbf{79.0}&$61.4$&$62.5$& $67.1$&$64.1$&$75.5$&$91.8$&$91.3$&$95.1$\\
ReLICv2 (ours) &\textbf{80.6}&$92.8$&$78.2$&\textbf{65.4}&\textbf{66.2}&\textbf{75.1}&\textbf{64.8}&\textbf{77.4}&\textbf{92.4}&$92.8$& $95.6$\\
\midrule
\multicolumn{12}{l}{\emph{Fine-tuned:}}\\
\midrule
Random Init \tiny{\citep{chen2020simple}} & $86.9$ & $95.9$ & $80.2$ & $76.1$  & $53.6$ & $91.4$ & $85.9$ & $64.8$ & $81.5$ & $72.6$ & $92.0$  \\
Supervised-IN \tiny{\citep{chen2020simple}} & $88.3$ & $97.5$ & \textbf{86.4} & $75.8$ & $64.3$ & $92.1$ & $86.0$ & $74.6$ & $92.1$ & $93.3$ & $97.6$ \\
SimCLR \tiny{\citep{chen2020simple}} & $88.2$ & $97.7$ & $85.9$ & $75.9$ & $63.5$ & $91.3$ & $88.1$ & $73.2$ & $89.2$ & $92.1$ & $97.0$  \\
BYOL \tiny{\citep{grill2020bootstrap}} &$88.5$&\textbf{97.8}&$86.1$&$76.3$&$63.7$&$91.6$&$88.1$ &$76.2$&$91.7$&\textbf{93.8}&$97.0$\\
ReLICv2 (ours) &\textbf{88.7}&$97.7$&$85.3$&\textbf{76.7}&\textbf{64.7}&\textbf{92.3}&\textbf{88.7}&\textbf{76.9}&\textbf{92.2}&$93.2$& \textbf{97.9}\\
\bottomrule
\end{tabular}
\end{adjustbox}
\end{center}
\caption{Accuracy (in \%) of transfer performance of a ResNet50 pretrained on ImageNet.}
\label{table.transfer}
\end{table}

\subsection{Robustness and OOD Generalization}  \label{app.robust_ood}

The robustness and out-of-distribution (OOD) generalization abilities of \alpharelic{} representations are tested on several detasets. 
We use ImageNetV2 \citep{recht2019imagenet} and ImageNet-C \citep{hendrycks2019benchmarking} datasets to evaluate robustness. 
ImageNetV2 \citep{recht2019imagenet} has three sets of $10000$ images that were collected to have a similar distribution to the original
ImageNet validation set, while ImageNet-C \citep{hendrycks2019benchmarking} consists of 15 synthetically generated corruptions (e.g. blur, noise) that are added to the ImageNet validation set. 

For OOD generalization we examine the performance on ImageNet-R \citep{hendrycks2021many}, ImageNetSketch \citep{wang2019learning} and ObjectNet \citep{barbu2019objectnet}. 
ImageNet-R \citep{hendrycks2021many} consists of $30000$ different renditions (e.g. paintings, cartoons) of $200$ ImageNet classes, while ImageNet-Sketch \citep{wang2019learning} consists of $50000$ images, $50$ for each ImageNet class, of object sketches in the black-and-white color scheme. 
These datasets aim to test robustness to different textures and other naturally occurring style changes and are out-of-distribution to the ImageNet training data. 
ObjectNet \citep{barbu2019objectnet} has $18574$ images from differing viewpoints and backgrounds compared to ImageNet.

On all datasets we evaluate the representations of a standard ResNet50 encoder under a linear evaluation protocol akin to Section \ref{sec:results:linear}, i.e. we freeze the pretrained representations and train a linear classifier using the labelled ImageNet training set; the test evaluation is performed zero-shot, i.e no training is done on the above datasets. 
As we have seen in Table \ref{table.imagenet_v2_c}, \alpharelic{} learns more robust representations and outperforms both the supervised baseline and the competing self-supervised methods on ImageNetV2 and ImageNet-C. We provide a detailed breakdown across the different ImageNet-C corruptions in Table \ref{table.imagenet_c_corruptions}. 
\begin{table*}[h]
\begin{center}
\begin{adjustbox}{max width=\textwidth}
\begin{tabular}{lccc|cccc|cccc|cccc}
\hline
& & & & \multicolumn{4}{|c|}{Blur} & \multicolumn{4}{|c|}{Weather} & \multicolumn{4}{|c}{Digital}  \\
Method & Gauss & Shot & Impulse  &  Defocus & Glass & Motion & Zoom & Snow & Frost & Fog & Bright & Contrast & Elastic & Pixel & JPEG  \\
\hline
\; Supervised {\tiny \citep{lim2019fast}} & 37.1 & 35.1 & 30.8 & 36.8 & \textbf{25.9} & 34.9 & \textbf{38.1}  & 34.5 & 40.7 & 56.9 & 68.1 & 40.6  & \textbf{45.6}  & 32.6  & 56.0  \\ 
\hline
\; SimCLR {\tiny\citep{chen2020simple}}  & 29.1 & 26.3 & 17.3 & 22.1 & 14.7 & 20.0 & 18.6 & 27.2  & 33.3 & 46.2 & 59.7 & 53.9 & 31.0 & 24.2 & 43.9  \\
\; BYOL {\tiny \citep{grill2020bootstrap}} & 41.5   & 38.7 & 31.9 & 37.8 & 22.5 & 31.6 & 29.6 & 35.1 & 42.9 & 60.1 & 69.0 & 58.4 & 41.5 & 46.3 & 55.9 \\
\; \relic{} {\tiny \citep{mitrovic2020representation}} & \textbf{43.4}  & \textbf{40.7}  & \textbf{36.6} & \textbf{40.5} & 24.5 & 34.3  & 30.5 & 36.6 & 43.8 & 61.4 & 69.5  &  59.5 & 42.8 & \textbf{46.8} & 57.3   \\
\; \alpharelic{} {\tiny (ours)} & 41.6  & 39.0 & 31.1 & 39.7  & 22.6  & \textbf{35.2} & 34.5 & \textbf{40.1} & \textbf{46.1} & \textbf{64.5} & \textbf{71.0} & \textbf{60.0} & 44.6 & 46.6 & \textbf{58.4}   \\ 
\hline
\end{tabular}
\end{adjustbox}
\end{center}
\caption{Top-1 accuracies for for Gauss, Shot, Impulse, Blur, Weather, and Digital
corruption types on ImageNet-C. \label{table.imagenet_c_corruptions}}
\end{table*}

\newpage
\section{Analysis}

\begin{wrapfigure}{r}{0.6\textwidth}
\vspace{-0.5cm}
    \centering
    \includegraphics[width=0.6\textwidth]{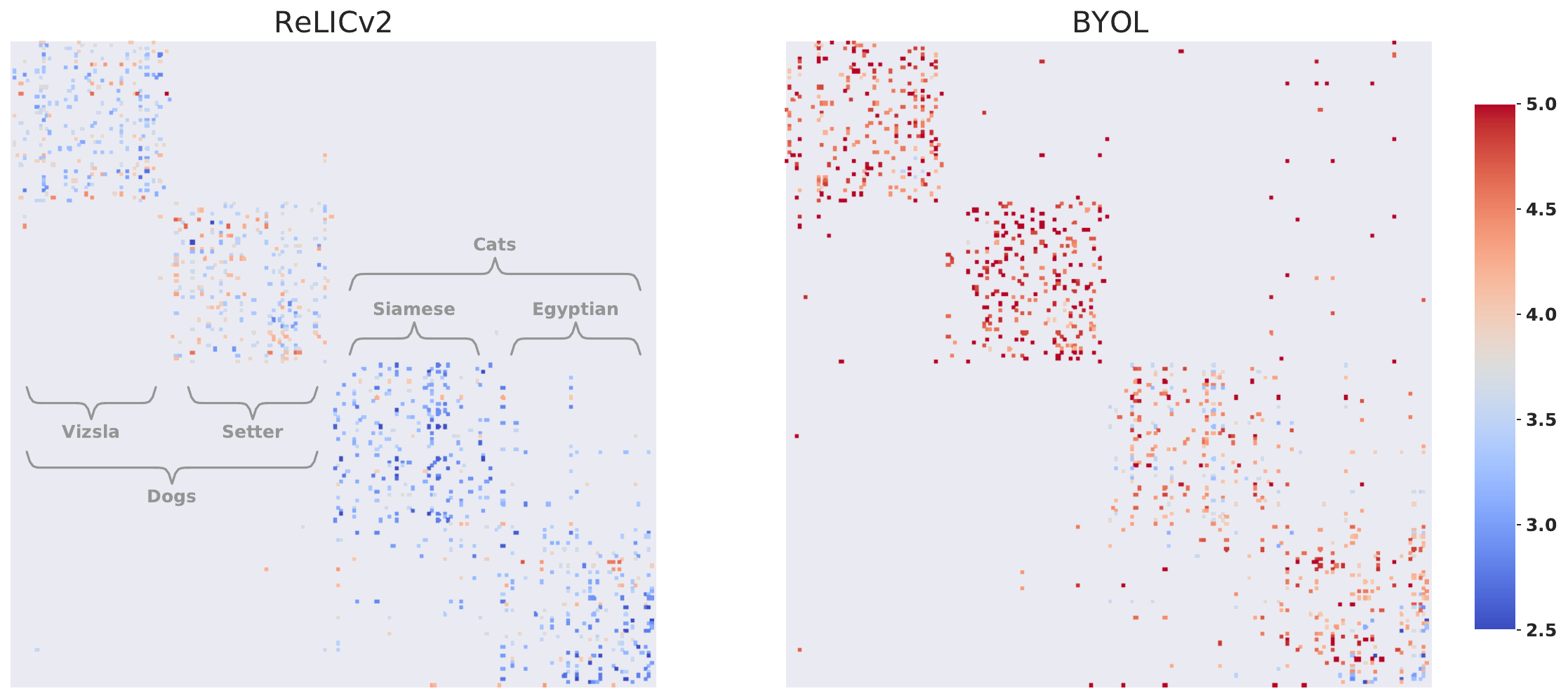}
  \caption{Distances between nearest-neighbour representations. Each coloured point in a row represents one of the five nearest neighbours of the representation of that image where the colour indicates the distance between the points. 
\label{fig:relic_heatmap}}
\vspace{-0.3cm}
\end{wrapfigure}

\paragraph{Class confusion.}
To understand the effect of the invariance term in \alpharelic{}, we look at the distances between learned representations of closely related classes. 
Figure~\ref{fig:relic_heatmap} illustrates the Euclidean distances between nearest-neighbour representations learned by \alpharelic{} and BYOL on ImageNet using the protocol described in section~\ref{sec:res}. 
Here we pick two breeds of dog and two breeds of cat. 
Each of these four classes has 50 points associated with it from the ImageNet validation set, ordered contiguously. 
Each row represents an image and each coloured point in a row represents one of the five nearest neighbours of the representation of that image where the colour indicates the distance between the image and the nearest neighbour. 
Representations which align perfectly with the underlying class structure would exhibit a perfect block-diagonal structure, i.e. their nearest neighbours all belong to the same underlying class. 
We see that \alpharelic{} learns representations whose nearest neighbours are closer and exhibit less confusion between classes and super-classes than BYOL.

\newpage
\section{Pretraining on Joint Foto Tree (JFT-300M) -- implementation details and additional results} \label{app.results_jft}

\subsection{Linear evaluation} 
We test how well \alpharelic{} scales to much larger datasets by pretraining representations using the Joint Foto Tree (JFT-300M) dataset which consists of 300 million images from more than 18k classes \citep{hinton2015distilling, chollet2017xception, sun2017revisiting}.
We then evaluate the learned representations on the ImageNet validation set under the same linear evaluation protocol as described in section \ref{sec:results:linear}. 
We compare \alpharelic{} against BYOL and Divide and Contrast (DnC) \citep{tian2021divide}, a method that was specifically designed to handle large and uncurated datasets and represents the current state-of-art in self-supervised JFT-300M pretraining. 
Table \ref{table.jft-top1} reports the top-1 accuracy when training the various methods using the standard ResNet50 architecture as the backbone for different number of ImageNet equivalent epochs on JFT-300M; implementation details can be found in the supplementary material. 
\alpharelic{} improves over DnC by more than $2\%$ when training on JFT for 1000 epochs and achieves better overall performance than competing methods while needing a smaller number of training epochs.

\begin{table}[h]
\begin{center}
\begin{adjustbox}{max width=\columnwidth}
\begin{tabular}{lcc}
\toprule
Method & Epochs & Top-1 \\
\toprule
\; BYOL \citep{grill2020bootstrap}  & 1000 &  67.0\\
\; Divide and Contrast \citep{tian2021divide} & 1000 & 67.9 \\
\; \alpharelic{} (ours) & 1000 & \textbf{70.3}  \\ 
\hline
\; BYOL \citep{grill2020bootstrap} & 3000 & 67.6\\
\; Divide and Contrast \citep{tian2021divide} & 3000 & 69.8 \\
\; \alpharelic{} (ours) & 3000 & \textbf{71.1}  \\ 
\hline
\; BYOL \citep{grill2020bootstrap} & 5000 & 67.9\\
\; Divide and Contrast \citep{tian2021divide} & 4500 & 70.7 \\
\; \alpharelic{} (ours) & 5000 & \textbf{71.4}  \\ 
\hline
\end{tabular}
\end{adjustbox}
\end{center}
\caption{Top-1 accuracy (in \%) on ImageNet when learning representations using the JFT-300M dataset. Each method is pre-trained on JFT-300M for an ImageNet-equivalent number of epochs and evaluted on the ImageNet validation set under a linear evaluation protocol.}
\label{table.jft-top1}
\end{table}

For results reported in Table \ref{table.jft-top1}, we use the following training and evaluation protocol.
To pretrain \alpharelic{} on the Joint Foto Tree (JFT-300M) dataset, we used a base learning rate of $0.3$ for pretraining the representations for $1000$ ImageNet-equivalent epochs. For longer pretraining of $3000$ and $5000$ ImageNet-equivalent epochs, we use a lower base learning rate of $0.2$. We set the target exponential moving average to $0.996$, the contrast scale to $0.3$, temperature to $0.2$ and the saliency mask apply probability to $0.15$ for all lenghts of pretraining.
For $1000$ and $5000$ ImageNet-equivalent epochs we use $2.0$ as the invariance scale, while for $3000$ ImageNet-equivalent epochs, we use invariance scale $1.0$. We then follow the linear evaluation protocol on ImageNet described in Appendix \ref{app.linear_eval}. We train a linear classifier on top of the pretrained representations from JFT-300M with stochastic  gradient descent with Nesterov momentum for 100 epochs using batch size of 256, learning rate of 0.5 and momentum of 0.9.

\subsection{Transfer}
We evaluate the transfer performance of JFT-300M pretrained representations under the linear evaluation protocol.
For this, we freeze the encoder and train only linear classifier on top of the frozen encoder output, i.e. representation.
As before in \ref{sec:supp:transfer}, we follow the transfer performance evaluation protocol as outlined in \citep{grill2020bootstrap,chen2020simple}. In line with prior work, for Food101~\citep{bossard2014food}, CIFAR10~\citep{krizhevsky2009learning}, CIFAR100~\citep{krizhevsky2009learning}, Birdsnap~\citep{berg2014birdsnap}, SUN397 (split 1)~\citep{xiao2010sun}, DTD (split 1)~\citep{cimpoi2014describing}, and Cars~\citep{krause20133d} we report the top-1 accuracy on the test set, and for Aircraft~\citep{maji2013fine}, Pets~\citep{parkhi2012cats}, Caltech101~\citep{fei2004learning}, and Flowers~\citep{nilsback2008automated} we report the mean per-class accuracy as the relevant metric in the comparisons. For DTD and SUN397, we only use the first split, of the 10 provided splits in the dataset.

We train on the training sets of the individual datasets and sweep over different values of the models hyperparameters. 
To select the best hyperparameters, we use the validation sets of the individual datasets.
Using the chosen hyperparameters, we train the linear layer from scratch using the merged training and validation data and test on the held out test data in order to obtain the numbers reported in Table \ref{table.transfer_jft}.
We swept over learning rates \{$.01$, $0.1$, $0.2$, $0.25$, $0.3$, $0.35$, $0.4$, $1.$, $2.$\}, batch sizes \{$128$, $256$, $512$, $1024$\}, weight decay between \{$1\mathrm{e}{-6}$, $1\mathrm{e}{-5}$, $1\mathrm{e}{-4}$, $1\mathrm{e}{-3}$, $0.01$, $0.1$\}, warmup epochs \{$0$, $10$\}, momentum \{$0.9$, $0.99$\}, Nesterov \{True, False\}, and the number of training epochs \{$60$, $80$, $100$\}.
Models were trained with the SGD optimizer with momentum.

As can be seen from Table \ref{table.transfer_jft}, longer pretraining benefits transfer performance of \alpharelic{}. 
Although DnC \citep{tian2021divide} was specifically developed to handle uncurated datasets such as JFT-300M, we see that \alpharelic{} has comparable performance to DnC in terms of the number of datasets with state-of-the-art performance among self-supervised representation learning methods; this showcases the generality of \alpharelic{}.

\begin{table}[h]
\begin{center}
\setlength\tabcolsep{2pt}
\begin{adjustbox}{max width=\columnwidth}
\begin{tabular}{lccccccccccc}
\midrule
Method & Food101 & CIFAR10 &  CIFAR100 &  Birdsnap &  SUN397  & Cars & Aircraft & DTD & Pets & Caltech101 & Flowers \\
\midrule
BYOL-5k \tiny{\citep{grill2020bootstrap}} & 73.3 & 89.8 & 72.4 & 38.2 & 61.8 & 64.4 & 54.4 & 75.5 & 77 & 90.1 & 94.3 \\
DnC-4.5k \tiny{\citep{tian2021divide}} & \textbf{78.7} & \textbf{91.7} & \textbf{74.9} & 42.1 & 65.0 & 75.3 & 54.1 & 76.6 & \textbf{86.1} & 90.2 & \textbf{98.2} \\
ReLICv2-1k (ours) & 77.5 & 90.2 & 72.6 & 47.4 & 64.5 & 74.4 & 62.9 & \textbf{77.0} & 84.9 & \textbf{92.2} & 94.5 \\
ReLICv2-5k (ours) & 78.3 & 89.9 & 73.0 & \textbf{49.4} & \textbf{65.6} & \textbf{76.9} & \textbf{65.5} & 76.8 & 85.1 & 91.4 & 95.7 \\
\bottomrule
\end{tabular}
\end{adjustbox}
\end{center}
\caption{Accuracy (in \%) of transfer performance of a ResNet50 pretrained on JFT under the linear transfer evaluation protocol. xk refers to the length of pretraining in ImageNet-equivalent epochs, e.g. 1k corresponds to 1000 ImageNet-equivalent epochs of pretraining.}
\label{table.transfer_jft}
\end{table}

\newpage
\subsection{Robustness and OOD Generalization} 
We also tested the robustness and out-of-distribution (OOD) generalization of \alpharelic{} representations pretrained on JFT.
We use the same set-up described in \ref{app.robust_ood} where we freeze the pretrained representations on JFT-300M, train a linear classifier using the labelled ImageNet training set and perform zeroshot test evaluation on datasets testing robustness and OOD generalization.
As in \ref{app.robust_ood}, we evaluated robustness using the ImageNetV2 \citep{recht2019imagenet} and ImageNet-C \citep{hendrycks2019benchmarking} datasets and OOD generalization using ImageNet-R \citep{hendrycks2021many}, ImageNetSketch \citep{wang2019learning} and ObjectNet \citep{barbu2019objectnet} datasets. 
We report the robustness results in Table \ref{table.jft_imagenet_v2} and the OOD generalization results in Table \ref{table.jft_ood}. 
We notice that \alpharelic{} representations pretrained on JFT-300M for different number of ImageNet-equivalent epochs have worse robustness and OOD generalization performance compared to \alpharelic{} representations pretrained directly on ImageNet (see Table \ref{table.imagenet_v2_c} for reference). 
Given that the above datasets have been specifically constructed to measure the robustness and OOD generalization abilities of models pretrained on ImageNet (as they have been constructed in relation to ImageNet), this result is not entirely surprising.
We hypothesize that this is due to there being a larger discrepancy between datasets and JFT-300M than these datasets and ImageNet and as such JFT-300M-pretrained representations perform worse than ImageNet-pretrained representations.
Additionally, note that pretraining on JFT-300M for longer does not necessarily result in better downstream performance on the robustness and out-of-distribution datasets. 

\begin{table}[!h]
\begin{subtable}{.5\linewidth}
\centering
\begin{tabular}{lcccc}
\toprule
Epochs & MF & T-0.7 & Ti & IN-C    \\
\midrule
1000 & 57.6 & 66.7  & 73.0 & 32.9  \\ 
3000   & 58.6 & 67.5 & 73.4 & 32.8 \\ 
5000  & 59.1 &  67.3 & 73.3 & 33.5 \\ 
\bottomrule
\end{tabular}
\caption{ImageNetv2 dataset.}
\label{table.jft_imagenet_v2}
\end{subtable}%
\begin{subtable}{.5\linewidth}
\centering
\begin{tabular}{lcccc}
\toprule
Epochs & IN-R & IN-Sketch &  ObjectNet    \\
\midrule
1000 & 20.4  & 6.7  & 20.3  \\ 
3000 &  20.3 & 8.7  & 21.3  \\
5000 & 20.3 & 5.4   & 20.9 \\
\bottomrule
\end{tabular}
\caption{ImageNet-R, ImageNet-Sketch and ObjectNet datasets.}
\label{table.jft_ood}
\end{subtable}
\caption{Top-1 Accuracy (in \%) under linear evaluation on the the ImageNet-R (IN-R), ImageNet-Sketch (IN-S) and ObjectNet out-of-distribution datasets and on ImageNetV2 dataset for \alpharelic{} pre-trained on JFT-300M for different numbers of ImageNet-equivalent epochs. We evaluate on all three variants on ImageNetV2 -- matched frequency (MF), Threshold 0.7 (T-0.7) and Top Images (TI). The results for ImageNet-C (IN-C) are averaged across the 15 different corruptions.}
\end{table}

\clearpage
\section{\alpharelic{} pseudo-code in Jax}

Listing \ref{alg:relicv2} provides PyTorch-like pseudo-code for \alpharelic{} detailing how we apply the saliency masking and how the different views of data are combined in the target network setting. We also provide a direct comparison with the pseudo-code for \relic{} \citep{mitrovic2020representation} highlighed in listing \ref{alg:relic}. Note that \texttt{loss\_relic} is computed using equation \ref{eq:loss_relic}. 

\noindent\begin{minipage}[t]{0.48\textwidth}
  \begin{algorithm}[H]
\begin{lstlisting}[language=Python, label={alg:relicv2}, caption=Pseudo-code for \alpharelic.]
'''
f_o: online network: encoder + comparison_net
g_t: target network: encoder + comparison_net
gamma: target EMA coefficient
n_e: number of negatives
p_m: mask apply probability 
'''
for x in batch: # load a batch of B samples 
    # Apply saliency mask and remove background
    x_m = remove_background(x) 
    for i in range(num_large_views):
        # Select either original or background-removed
        # Image with probability p_m
        x = Bernoulli(p_m) ? x_m : x
        # Do large random view and augment
        xl_i = aug(crop_l(x)) 
        
        ol_i = f_o(xl_i) 
        tl_i = g_t(xl_i)
        
    for i in range(num_small_views):
        # Do small random view and augment
        xs_i = aug(crop_s(x))
        # Small views only go through the online network
        os_i = f_o(xs_i) 
     
    loss = 0  
    # Compute loss between all pairs of large views
    for i in range(num_large_views): 
        for j in range(num_large_views):
            loss += loss_relic(ol_i, tl_j, n_e)
            
    # Compute loss between small views and large views
    for i in range(num_small_views):
        for j in range(num_large_views):
            loss += loss_relic(os_i, tl_j, n_e)
    scale = (num_large_views + num_small_views) * 
            num_large_views
    loss /= scale
    
    # Compute grads, update online and target networks
    loss.backward()
    update(f_o)
    g_t = gamma * g_t + (1 - gamma) * f_o 
    
\end{lstlisting}
\end{algorithm}

\end{minipage}%
\hfill%
\begin{minipage}[t]{0.48\textwidth}
 \begin{algorithm}[H]

\begin{lstlisting}[language=Python, label={alg:relic}, caption=Pseudo-code for \relic.]
'''
f_o: online network: encoder + comparison_net
g_t: target network: encoder + comparison_net
gamma: target EMA coefficient
n_e: number of negatives
'''
# load a batch of B samples 
for x in batch:
    # Apply augmentations
    x1 = aug(x)
    x2 = aug(x)
        
    o1, o2 = f_o(x1), f_o(x2)
    t1, t2 = f_t(x1), f_t(x2)
    
    # Compute loss between augmented views
    loss  = loss_relic(o1, t2, n_e) +      
            loss_relic(o2, t1, n_e) 
    loss /= 2
     
    loss.backward()
     
    # Compute grads, update online and target networks
    loss.backward()
    update(f_o)
    g_t = gamma * g_t + (1 - gamma) * f_o 
\end{lstlisting}

\end{algorithm}

\end{minipage}

\newpage
\section{Image Preprocessing}  \label{app.image_augmentations}

\subsection{Augmentations}

Following the data augmentations protocols of \citep{chen2020simple, grill2020bootstrap, caron2020unsupervised}, \alpharelic{} uses a set of augmentations to generate different views of the original image which has three channels, red $r$, green $g$ and blue $b$ with $r,g,b \in [0,1]$.

The augmentations used, in particular (corresponding to \texttt{aug} in Listing~\ref{alg:relic}) are the same as in \citep{grill2020bootstrap} and are generated as follows; for exact augmentations parameters see Table~\ref{table:aug}). 
The following sequence of operations is performed in the given order.
\begin{enumerate}
    \item Crop the image: Randomly select a patch of the image, between a minimum and maximum crop area of the image, with aspect ratio sampled log-uniformly in $[3/4, 4/3]$. Upscale the patch, via bicubic interpolation, to a square image of size $s \times s$.
    \item Flip the image horizontally.
    \item Colour jitter: randomly adjust brightness, contrast, saturation and hue of the image, in a random order, uniformly by a value in $[-a, a]$ where $a$ is the maximum adjustment (specified below).
    \item Grayscale the image, such that the  channels are combined into one channel with value $0.2989r + 0.5870g + 0.1140b$.
    \item Randomly blur. Apply a $23\times 23$ Gaussian kernel with standard deviation sampled uniformly in $[0.1, 2.0]$.
    \item Randomly solarize: threshold each channel value such that all values less than $0.5$ are replaced by $0$ and all values above or equal to $0.5$ are replaced with $1$.
\end{enumerate}
Apart from the initial step of image cropping, each step is executed with some probability to generate the final augmented image. These probabilities and other parameters are given in Table~\ref{table:aug}, separately for augmenting the original image $x_i$ and the positives $\mathcal{P}(x_i)$. Note that we use 4 large views of size $224\times 224$ pixels and 2 small views of $96\times 96$ pixels; to get the first and third large views and the first small view we use the parameters listed below for odd views, while for the second and fourth large view and the second small view we use the parameters for even views.

\begin{table}[h]
\begin{center}
\begin{tabular}{lcc}
\midrule
\textbf{Parameter} & Even views & Odd views \\
\midrule
Probability of randomly cropping & 50\% & 50\% \\
Probability of horizontal flip & 50\% & 50\% \\
Probability of colour jittering & 80\% & 80\% \\
Probability of grayscaling & 20\% & 20\% \\
Probability of blurring & 100\% & 10\% \\
Probability of solarization & 0\% & 20\% \\
Maximum adjustment $a$ of brightness & 0.4 & 0.4 \\
Maximum adjustment $a$ of contrast & 0.4 & 0.4 \\
Maximum adjustment $a$ of saturation & 0.2 & 0.2 \\
Maximum adjustment $a$ of hue & 0.1 & 0.1 \\
Crop size $s$ & 224 & 96 (small), 224 (large) \\
Crop minimum area & 8\% & 5\% (small), 14\% (large) \\
Crop maximum area & 100\% & 14\% (small), 100\% (large) \\
\bottomrule
\end{tabular}
\end{center}
\caption{Parameters of data augmentation scheme. Small/large indicates small or large crop.}
\label{table:aug}
\end{table}

\subsection{Saliency Masking}  \label{supp:saliency_masking}

Using unsupervised saliency masking enables us to create positives for the anchor image with the background largely removed and thus the learning process will rely less on the background to form representations. 
This encourages the representation to localize the objects in the image \citep{zhao2020distilling}.

\begin{figure}[h]
    \centering
\includegraphics[width=0.6\textwidth]{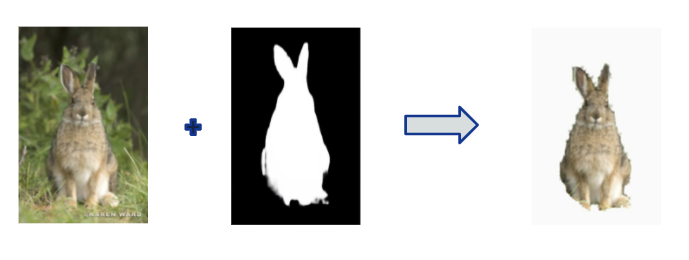}
  \caption{Illustration of how for each image in the ImageNet training set (left) we use our unsupervised version of DeepUSPS to obtain the saliency mask (middle) which we then apply on top of the image to obtain the image with the background removed (right).
\label{fig:saliency_viz}}
\end{figure}

We develop a fully unsupervised saliency estimation method that uses the self-supervised refinement mechanism from DeepUSPS \citep{deepusps} to compute saliency masks for each image in the ImageNet training set. 
By applying the saliency masks on top of the large views, we obtain masked images with the background removed. 
To further increase the background variability, instead of using a black background for the images, we apply a homogeneous grayscale to the background with the grayscale level randomly sampled for each image during training. 
We also use a foreground threshold such that we apply the saliency mask only if it covers at least $5\%$ of the image.
The masked images with the grayscaled background are used only during training. 
Specifically, with a small probability $p_m$ we selected the masked image of the large view in place of the large view.
Figure \ref{fig:saliency_viz} shows how the saliency masks are added on top of the images to obtain the images with grayscale background.

\vspace{-0.2cm}
\subsubsection{Training the saliency detection network to obtain saliency masks} \label{apx:saliency_masking_deepusps}

DeepUSPS \citep{deepusps} is a  saliency prediction method that uses self-supervision to refine pseudo-labels from a number of handcrafted saliency methods.
To obtain saliency masks for the images in ImageNet, we build a new saliency detection method that leverages the self-supervised refinement mechanism from DeepUSPS \citep{deepusps}. To this end, we firstly sample a random subset of $2500$ ImageNet images; note that the original implementation of DeepUSPS uses $2500$ images from the MSRA-B dataset. We instead use a randomly selected subset of the ImageNet training set of the same size to ensure a fair comparison to previous work. We compute initial saliency masks for the $2500$ ImageNet images using the following handcrafted methods: Robust Background Detection (RBD) \citep{zhu2014saliency}, Manifold Ranking (MR) \citep{yang2013saliency}, Dense and Sparse Reconstruction (DSR) \citep{li2013saliency} and Markov Chain (MC) \citep{jiang2013saliency}. Note that these methods do not make use of any supervised label information. 

We then follow the two-stage mechanism proposed by DeepUSPS \citep{deepusps} to obtain a saliency prediction network.  
In the first stage, the noisy pseudo-labels from each handcrafted method are iteratively refined. 
In the second stage, these refined labels from each handcrafted saliency method are used to train the final saliency detection network. The saliency detection network is then used to compute the saliency masks for all images in the ImageNet training set. 
For the refinement procedure and for training the saliency detection network, we adapt the publicly available code for training DeepUSPS: \url{https://tinyurl.com/wtlhgo3}.
Note that the official implementation for DeepUSPS uses as backbone a DRN-network \citep{yu2017dilated} which was pretrained on CityScapes \citep{Cordts2016Cityscapes} with supervised labels. 
To be consistent with our fully-unsupervised setting, we replace this network with a ResNet50 2x model which was pretrained on ImageNet using the self-supervised objective from SWaV \citep{caron2020unsupervised}. 
We used the publicly available pretrained SWaV model from: \url{https://github.com/facebookresearch/swav}. 

To account for this change in the architecture, we adjust some of the hyperparameters needed for the the two-stage mechanism of DeepUSPS. 
In the first stage, the pseudo-generation networks used for refining the noisy pseudo-labels from each of the handcrafted methods are trained for $25$ epochs in three self-supervised iterations. We start with a learning rate of $1e-5$ which is doubled during each iteration. In the second stage, the saliency detection network is trained for $200$ epochs using a learning rate of $1e-5$. We use the Adam optimizer with momentum set to $0.9$ and a batch size of $10$. The remaining hyperparameters are set in the same way as they are in the original DeepUSPS code.

Next, for different probabilities $p_m$ of removing the background of the large augmented views during training, we report below the top-1 accuracy under linear evaluation on ImageNet.
\vspace{-0.2cm}
\begin{center}
\begin{tabular}{l|ccccc}
 $p_m$ & 0.0 & 0.1 & 0.15 & 0.2 & 0.25 \\
\hline
Top-1 & 76.8 & \textbf{77.1} & 76.8 & 76.8 & 76.7 \\
\end{tabular}
\vspace{-0.1cm}
\end{center}
Applying the saliency masks 10\% of the time results in the best performance and significantly improves over not using masking ($p_m=0$).

\newpage
\section{Ablations} 
In order to determine the sensitivity of \alpharelic{} to different model hyperparameters, we perform an extensive ablation study. 
Unless otherwise noted, in this section we report results after $300$ epochs of pretraining.
As saliency masking is one of the main additions of \alpharelic{} on top of \relic{} and was not covered extensively in the main text, we start our ablation analysis with looking into the effect of different modelling choices for it.

\subsection{Using different datasets for obtaining the saliency masks}

In the main text in Sections \ref{sec:results:linear}, \ref{sec:res:transfer}, \ref{sec:res:robust} we used a saliency detection network trained only on a randomly selected subset of $2500$ ImageNet images using the refinement mechanism proposed by DeepUSPS \citep{deepusps}.
Here we explore whether using additional data could help improve the performance of the saliency estimation and of the overall representations learnt by \alpharelic{}. 
For this purpose, we use the MSRA-B dataset \citep{liu2010learning}, which was originally used by DeepUSPS to train their saliency detection network. 
MSRA-B consists of $2500$ training images for which handcrafted masks computed with the methods Robust Background Detection (RBD) \citep{zhu2014saliency}, Hierarchy-associated Rich Features (HS) \citep{zou2015harf}, Dense and Sparse Reconstruction (DSR) \citep{li2013saliency} and Markov Chain (MC) \citep{jiang2013saliency} are already available. 
We use the same hyperparameters as described in Section \ref{apx:saliency_masking_deepusps} to train our saliency detection network on MSRA-B; note that these are the same hyperparameters we use for training our saliency detection network on ImageNet. 

We explored whether using saliency masks obtained from training the saliency detection network on the MSRA-B affects performance of \alpharelic{} pre-training on ImageNet. 
We noticed that for \alpharelic{} representations pretrained on ImageNet for $1000$ epochs, we get $77.2\%$ top-1 and $93.3\%$ top-5 accuracy under linear evaluation on the ImageNet validation set for a ResNet50 (1x) encoder. 
The slight performance gains may due to the larger variety of images in MSRA-B used for training the saliency detection network, as opposed to the random sample of $2500$ ImageNet images that we used for training the saliency detection network directly on the ImageNet dataset. 

We also explored training the saliency detection network on $5000$ randomly selected images from the ImageNet dataset and this resulted in the model overfitting, which degraded the quality of the saliency masks and resulted in a \alpharelic{} performance of $76.7\%$ top-1 and $93.3\%$ top-5 accuracy on the ImageNet validation set after $1000$ epochs of pretraining on ImageNet training set.

The results for \alpharelic{} in Section \ref{app.results_jft} are obtained by applying the saliency detection network trained on MSRA-B to all images in JFT-300M and then applying the saliency masks to the large augmented views during training as described in Section \ref{supp:saliency_masking}.

\subsection{Analysis and ablations for saliency masks}

Using saliency masking during \alpharelic{} training enables us to learn representations that focus on the semantically-relevant parts of the image, i.e. the foreground objects, and as such the learned representations should be more robust to background changes. 
We investigate the impact of using saliency masks with competing self-supervised benchmarks, the effect of the probability $p_m$ of applying the saliency mask to each large augmented view during training as well as the robustness of \alpharelic{} to random masks and mask corruptions. 
For the ablation experiments described in this section, we train the models for $300$ epochs. 

\paragraph{Using saliency masks with competing self-supervied methods.} We evaluate the impact of using saliency masks with competing self-supervised methods such as BYOL \citep{grill2020bootstrap}. 
This method only uses two large augmentented views during training and we randomly apply the saliency masks, in a similar way as described in Section \ref{supp:saliency_masking}, to each large augmented view with probability $p_m$. 
We report in Table \ref{table.ablation_byol_simclr_masks} the top-1 and top-5 accuracy under linear evaluation on ImageNet for different settings of $p_m$ for removing the background of the augmented images. 
We notice that saliency masking also helps to improve performance of BYOL.

\begin{table}[h]
\begin{center}
\begin{tabular}{llcccccc}
\toprule
& Mask probability $p_m$ & 0 & 0.1 & 0.15 & 0.2 & 0.25 & 0.3  \\
\midrule
BYOL& Top-1 & 73.1 & 73.4 & 73.2 & 73.3 & 72.8  & 71.8 \\ 
& Top-5 & 91.2 & 91.3 & 91.2 & 91.3 & 90.8 & 90.1 \\
\bottomrule
\end{tabular}
\end{center}
\caption{Top-1 and top-5 accuracy (in \%) under linear evaluation on the ImageNet validation set for BYOL trained using different probabilities of using the saliency mask to remove the background of the augmented images. Models are trained for 300 epochs.}
\label{table.ablation_byol_simclr_masks}
\end{table}

\paragraph{Mask apply probability.} We also investigate the effect of using probabilities ranging from 0 to 1 for applying the saliency mask during training for \alpharelic{}. 
In addition, we explore further the effect of using different datasets for training the saliency detection network that is subsequently used for computing the saliency masks. Table \ref{table.ablation_mask_prob_imagenet_msrab} reports the top-1 and top-5 accuracy for varying the mask apply probability $p_m$ between 0 and 1 and for using the ImageNet vs. the MSRA-B dataset \citep{liu2010learning} for training our saliency detection network. 
Note that using the additional images from the MSRA-B dataset to train the saliency detection network results in better saliency masks which translates to better performance when using the saliency masks during \alpharelic{} training. 

\begin{table}[!h]
\begin{center}
\begin{tabular}{l|cc|cc}
\hline
& \multicolumn{2}{c}{Sal. network trained on ImageNet} & \multicolumn{2}{|c}{Sal. network trained on MSRA-B} \\
Mask probability $p_m$ & Top-1 & Top-5 & Top-1 & Top-5 \\
\hline
\; 0 & 75.2 & 92.4 & 75.2 & 92.4  \\
\; 0.05 & 75.3 & 92.6 & 75.2 & 92.6  \\
\; 0.1 & 75.4 & 92.5 & 75.3 & 92.4 \\
\; 0.15 & 75.2 & 92.5 & 75.5 & 92.5 \\
\; 0.2 & 75.2 & 92.5 & 75.6 & 92.6 \\
\; 0.25 & 75.0 & 92.3 & 75.3 & 92.5 \\ 
\; 0.3 & 75.1 & 92.3 &  74.8 & 92.4 \\ 
\; 0.4 & 75.0 & 92.3 &  75.3 & 92.5 \\
\; 0.5 & 74.7 & 92.2 & 75.0 & 92.4 \\
\; 0.6 & 75.0 &  92.3 & 75.0 & 92.3 \\
\; 0.7 & 74.4 & 92.3 & 74.6  & 92.0 \\
\; 0.8 & 73.9 & 91.7 & 75.0  & 92.1 \\
\; 0.9 & 74.0 & 91.7 & 74.6  & 92.0 \\
\; 1.0 & 73.7 & 91.7 & 74.5 & 92.0 \\
\hline
\end{tabular}
\end{center}
\caption{Top-1 and top-5 accuracy (in \%) under linear evaluation on the ImageNet validation set for a ResNet50 (1x) encoder set for different probabilities $p_m$ of using the saliency mask to remove the  background of the large augmented views during training  and for using different datasets to train the saliency detection network for computing the saliency masks. Models are trained for 300 epochs.}
\label{table.ablation_mask_prob_imagenet_msrab}
\end{table}

\paragraph{Random masks and mask corruptions.} 
To understand how important having accurate saliency masks for the downstream performance of representations is we also investigated using random masks, corrupting the saliency masks obtained from our saliency detection network and using a bounding box around the saliency masks during \alpharelic{} training. 

We explored using completely random masks, setting the saliency mask to be a random rectangle of the image and also a centered rectangle. 
As ImageNet images generally consists of images with objects centered in the middle of the image, we expect that using a random rectangle that is centered around the middle will cover a reasonable portion of the object. 
Table \ref{table.ablation_random_masks} reports the performance under linear evaluation on the ImageNet validation set when varying the size of the random masks to cover different percentage areas $a_p$ of the full image. 
We notice that improving the quality of the masks, by using random rectangle patches instead of completely random points in the image as the mask, results in better performance. 
However, the performance with random masks is $>1\%$ lower than using saliency masks from our saliency detection network. 
As expected, using centered rectangles instead of randomly positioned rectangles as masks results in better peformance.

\begin{table}[!h]
\begin{center}
\begin{tabular}{l|cc|cc|cc}
\hline
& \multicolumn{2}{c|}{Random} & \multicolumn{2}{c|}{Rectangle} & \multicolumn{2}{c}{Centered Rectangle} \\
Image percentage area $a_p$ & Top-1 & Top-5 & Top-1 & Top-5 & Top-1 & Top-5 \\
\hline
\; 10\% & 70.8 & 89.9 & 70.9 & 90.3 & 71.3 & 90.1 \\
\; 20\% & 72.2 & 90.7 & 73.1 & 91.3 & 73.4 & 91.3 \\
\; 30\% & 72.9 & 91.3 & 73.8 & 91.8 & 73.8 & 91.9 \\ 
\; 40\% & 73.1 & 91.4 & 74.2 & 91.9 & 74.1 & 92.0 \\
\; 50\% & 73.3 & 91.5 & 74.0 & 92.0 & 74.3 & 92.0 \\
\; 60\% & 73.6 & 91.8 & 74.2 & 92.1 & 74.3 & 92.2 \\
\; 70\% & 73.7 & 91.9 & 74.4 & 92.1 & 74.4 & 92.2 \\
\; 80\% & 74.1 & 92.1 & 74.4 & 92.2 & 74.2 & 92.1 \\
\; 90\% & 74.1 & 92.2 & 74.4 & 92.1 & 74.2 & 92.2 \\
\hline
\end{tabular}
\end{center}
\caption{Top-1 and top-5 accuracy (in \%) under linear evaluation on the ImageNet validation set for a ResNet50 (1x) encoder set for using different types of random masks that cover various  percentage areas ($a_p$) of the full image. These random masks are applied on top of the large augmented views during training with probability 0.1. Models are trained for 300 epochs.}
\label{table.ablation_random_masks}
\end{table}

Moreover, to test the robustness of \alpharelic{} to corruptions of the saliency masks, we add/remove from the masks a rectangle proportional to the area of the saliency mask. 
The mask rectangle is added/removed from the image center. 
Table \ref{table.ablation_mask_corruptions} reports the results when varying the area of the rectangle to be added/removed to cover different percentages $m_p$ of the saliency masks. 
We notice that while \alpharelic{} is robust to small corruptions of the saliency mask its performance drops in line with the quality of the saliency masks degrading. 

\begin{table}[!h]
\begin{center}
\begin{tabular}{l|cc|cc}
\hline
& \multicolumn{2}{c|}{Add rectangle to mask} & \multicolumn{2}{c}{Remove rectangle from mask} \\
Mask percentage area $m_p$ & Top-1 & Top-5 & Top-1 & Top-5  \\
\hline
\; 10\% & 75.2 & 92.5 & 75.2 & 92.3 \\
\; 20\% & 75.3 & 92.6 &  75.1 & 92.4 \\
\; 30\% & 75.1  & 92.3 &  74.7 & 92.2 \\ 
\; 40\% & 74.9  & 92.2 &  74.6 & 92.2 \\
\; 50\% & 74.9 & 92.4  &  74.5 & 92.0 \\
\; 60\% & 74.9 & 92.2 &  74.0 & 91.7 \\
\; 70\% & 74.8 & 92.2 & 73.6 & 91.7 \\
\; 80\% & 74.8 & 92.4 & 73.4 & 91.4 \\
\; 90\% & 74.7 & 92.2  & 73.0 & 91.3 \\
\; 100\% & 74.6 & 92.3 & 72.6 & 90.9 \\
\hline
\end{tabular}
\end{center}
\caption{Top-1 and top-5 accuracy (in \%) under linear evaluation on the ImageNet validation set for a ResNet50 (1x) encoder set for corrupting the saliency masks by adding/remove a rectangle from the image center. The rectangle is a percentage ($m_p$) of the saliency mask area (the higher the percentage the higher the corruption). The corrupted saliency masks are applied on top of the large augmented views during training with probability 0.1.}
\label{table.ablation_mask_corruptions}
\end{table}

Finally, we also explore corrupting the masks using a bounding box around the saliency mask which results in $74.5\%$ top-1 and $92.2\%$ top-5 accuracy under linear evaluation on the ImageNet validation set for a ResNet50 (1x) encoder trained for $300$ epochs with mask apply probability of $0.1$ Note that this performance is comparable to using random rectangles to mask the large augmented views during training (see Table \ref{table.ablation_random_masks}) and  is lower than directly using the saliency masks from the trained saliency detection network.

\subsection{Other model hyperparameters}

Now we turn our attention to ablating the effect of other model hyperparameters on the downstream performance of \alpharelic{} representations.
Note that these hyperparameters have been introduced and extensively ablated in prior work \citep{grill2020bootstrap,mitrovic2020representation,mitrovic2020less}.

\paragraph{Number of negatives.} 
As mentioned in Section~\ref{sec:method} \alpharelic{} selects negatives by randomly subsampling the minibatch in order to avoid false negatives. 
We investigate the effect of changing number of negatives in Table~\ref{table.ablation_negatives}.
We can see that the best performance can be achieved with relatively low numbers of negatives, i.e. just 10 negatives.
Furthermore, we see that using the whole batch as negatives has one of the lowest performances.

In further experiments, we observed that for longer pretraining (e.g. $1000$ epochs) there is less variation in performance than for pretraining for $300$ epoch which itself is also quite low.

\begin{table}[h]
\begin{center}
\begin{tabular}{lccc}
\hline
Number of negatives & Top-1 & Top-5 \\
\hline
\; 1 & 75.1 & 92.4  \\ 
\; 5 & 75.2 & 92.6 \\
\; 10 & 75.4 & 92.5  \\
\; 20 & 75.3 & 92.7  \\
\; 50 & 75.5 & 92.5 \\ 
\; 100 & 75.4 & 92.5 \\ 
\; 500 & 75.1 & 92.4 \\
\; 1000 & 75.3 & 92.6 \\
\; 2000 & 75.4 & 92.5 \\
\; 4096 & 75.2 & 92.6  \\
\hline
\end{tabular}
\end{center}
\caption{Top-1 and top-5 accuracy (in \%) under linear evaluation on the ImageNet validation set for a ResNet50 (1x) encoder set for different numbers of randomly selected negatives.  All settings are trained for $300$ epochs. }
\label{table.ablation_negatives}
\end{table}

\paragraph{Target EMA.} 
\alpharelic{} uses a target network whose weights are an exponential moving average (EMA) of the online encoder network which is trained normally using stochastic gradient descent; this is a setup first introduced in \citep{grill2020bootstrap} and subsequently used in \citep{mitrovic2020representation} among others.
The target network weights at iteration $t$ are $\xi_t = \gamma\xi_{t-1} + (1-\gamma) \theta_t$ where $\gamma$ is the EMA parameter which controls the stability of the target network ($\gamma=0$ sets $\xi_t=\theta_t$); $\theta_t$ are the parameters of the online encoder at time $t$, while $\xi_t$ are the parameters of the target encoder at time $t$. 
As can be seen from Table \ref{table.ablation_target_ema}, all decay rates between $0.9$ and $0.996$ yield similar performance for top-1 accuracy on the ImageNet validation set after pretraining for $300$ epochs indicating that \alpharelic{} is robust to choice of $\gamma$ in that range.
For values of $\gamma$ of $0.999$ and higher, the performance quickly degrades indicating that the updating of the target network is too slow.
Note that contrary to \citep{grill2020bootstrap} where top-1 accuracy drops below $20\%$ for $\gamma=1$, \alpharelic{} is significantly more robust to this setting achieving double that accuracy.

\begin{table}[h]
\begin{center}
\begin{tabular}{lccc}
\hline
$\gamma$ &Top-1 & Top-5 \\
\hline
\; 0 & 73.5 & 91.5 \\
\; 0.9 & 74.6 & 92.2  \\
\; 0.99 & 75.5 & 92.6 \\
\; 0.993   & 75.4 & 92.5  \\
\; 0.996  & 74.4 & 92.0  \\
\; 0.999 & 70.5 & 89.8 \\ 
\; 1.0 & 39.6 & 63.6 \\
\hline
\end{tabular}
\end{center}
\vspace{-0.4cm}
\caption{Top-1 and top-5 accuracy (in \%) under linear evaluation on the ImageNet validation set for a ResNet50 (1x) encoder set for different setting of the target exponentially moving average (EMA). All settings are trained for 300 epochs. }
\label{table.ablation_target_ema}
\end{table}

\end{document}